\documentclass{article}

\usepackage[preprint]{neurips_2026}

\usepackage[utf8]{inputenc} 
\usepackage[T1]{fontenc}    
\usepackage{hyperref}       
\usepackage{url}            
\usepackage{booktabs}       
\usepackage{amsfonts}       
\usepackage{nicefrac}       
\usepackage{microtype}      
\usepackage{xcolor}         
\usepackage{csquotes}
\usepackage{algorithmicx}
\usepackage{subcaption}

\usepackage{graphicx}
\usepackage{booktabs}
\usepackage{booktabs}
\usepackage{tabularx}
\usepackage{algorithm}
\usepackage{algpseudocode}
\usepackage{amsmath}
\usepackage{amssymb}
\usepackage{mathtools}
\usepackage{amsthm}
\newtheorem{condition}{Condition}
\usepackage{makecell}

\usepackage{multirow}
\usepackage{booktabs} 
\usepackage{array} 

\theoremstyle{plain}
\newtheorem{theorem}{Theorem}[section]
\newtheorem{proposition}[theorem]{Proposition}
\newtheorem{lemma}[theorem]{Lemma}

\theoremstyle{definition}
\newtheorem{definition}[theorem]{Definition}

\theoremstyle{remark}

\usepackage{cleveref}
\def\paperabb{TaSS}
\title{Learning a Task-Adaptive Low-Dimensional Semantic Space for Improved Visual Classification}

\author{
  Nikalal Kaldera \quad
  Wanqing Li \quad
  Chris Bunn \quad
  Philip O.~Ogunbona \\
  School of Computing and Information Technology \\
  University of Wollongong \\
  Wollongong, NSW 2500, Australia
}

\begin{document}

\maketitle
\begin{abstract}
Semantic visual classification commonly aligns visual embeddings with semantic spaces induced by pretrained text encoders. However, such spaces are typically high-dimensional, weakly aligned with visual feature geometry, and insufficiently discriminative for downstream classification tasks. We propose \paperabb{}, a framework for learning a task-adaptive low-dimensional semantic space that preserves semantic structure while explicitly optimizing class separability. \paperabb{} is constructed through a parameterized geometry transformation of pretrained text embeddings that enlarges inter-class margins via distance reshaping and improves intra-class compactness through prototype-preserving semantic constraints. The resulting semantic space maintains meaningful semantic relationships while producing a discriminative geometry tailored for classification. Frozen visual encoder features are subsequently aligned to \paperabb{} using a lightweight MLP, enabling improved discriminative learning without fine-tuning pretrained backbones. We further provide theoretical analysis showing that \paperabb{} preserves semantic consistency while enforcing enhanced inter-class separation and controlled intra-class variance. Extensive experiments on skeleton-based human action recognition, video classification, and image classification benchmarks demonstrate consistent improvements over existing semantic alignment approaches, with gains exceeding 3 percentage points on average and up to 8.5 points on challenging datasets.
\end{abstract}

\section{Introduction}

Semantic visual classification commonly relies on aligning visual representations with semantic embeddings derived from pretrained language models such as CLIP~\cite{clip}. In this paradigm, semantic embeddings encode relationships between classes independently of visual observations and provide transferable supervision for downstream recognition tasks. Existing approaches typically map visual features into a semantic space induced directly by pretrained text encoders (Fig.~1a).

Despite their strong generalization ability, pretrained semantic spaces are not necessarily optimal for discriminative visual classification. First, they are often high-dimensional and geometrically mismatched with compact visual representations extracted by modern vision encoders. Second, semantic proximity in language space does not always correspond to discriminative separability for visual recognition, causing semantically related classes to overlap. Third, directly inheriting the geometry of pretrained language embeddings provides limited control over intra-class structure and inter-class margins in downstream tasks~\cite{schall2025comprehensive}. Consequently, semantic alignment may preserve linguistic relationships while failing to produce sufficiently discriminative representations for classification.

This motivates the following question:
\textbf{\textit{Can we learn a semantic space that simultaneously preserves semantic relationships and optimizes discriminative geometry for visual classification?}}
More specifically, we seek a semantic space satisfying three properties: (a) \textit{dimensional compatibility with visual encoder features}, (b) \textit{enhanced inter-class separation and intra-class compactness}, and (c) \textit{preservation of the semantic structure inherited from pretrained text embeddings}.

\begin{figure*}[t]
\centering

\includegraphics[width=\textwidth,height=0.30\textheight,keepaspectratio]{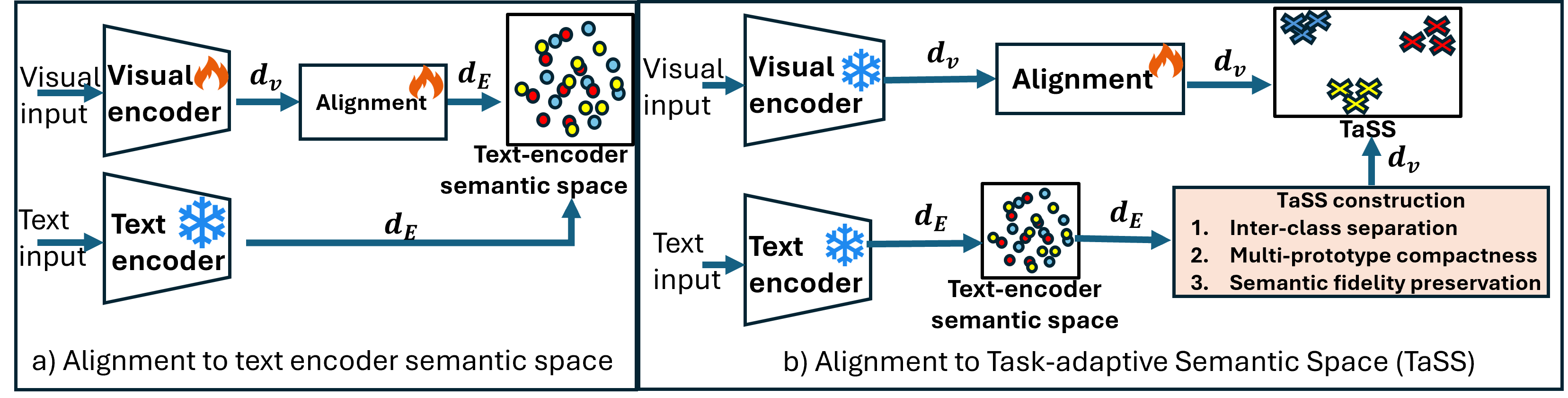}

\vspace{-0.5em}
\caption{\small Overview of text-guided visual classification. 
(a) Alignment to the original text-encoder semantic space that is dimensionally mismatched with visual features is not optimized for discrimination, limiting class separability. 
(b) Our method aligns frozen visual features to \paperabb{} constructed from the text-encoder space, with controllable dimensionality, enhanced inter-class separation, multi-prototype compactness, and semantic preservation.}

\label{fig:dss_framework}


\begin{minipage}{0.92\textwidth}
\centering
\captionof{table}{Comparison of dimensionality reduction methods.}
\vspace{-0.4em}
\setlength{\tabcolsep}{4pt}
\renewcommand{\arraystretch}{0.95}
\begin{tabular}{|l|c|c|c|c|c|}
\hline
\textbf{Method} & \textbf{Dim. Control} & \textbf{Separation} & \textbf{Compactness} & \textbf{Semantic Fidelity} & \textbf{Supervised} \\
\hline
MDS & High & Low & Low & Yes & No \\
\hline
t-SNE & High & Low & Medium & No & No \\
\hline
UMAP & High & Medium & Medium & No & No \\
\hline
LDA & Low & High & High & No & Yes \\
\hline
\paperabb{} (Ours) & High & High & High & Yes & Yes \\
\hline
\end{tabular}
\label{tab:dr_methods}
\end{minipage}

\vspace{-0.8em}
\end{figure*}

Conventional dimensionality reduction methods only partially address these objectives. While they may reduce dimensional mismatch, they generally lack mechanisms for jointly preserving semantic consistency and enforcing discriminative class geometry, as summarized in Table~\ref{tab:dr_methods}.

To address this limitation, we propose \paperabb{}, a framework for learning a discriminative low-dimensional semantic space tailored for visual classification. \paperabb{} is constructed through a parameterized transformation of pretrained semantic embeddings that explicitly reshapes semantic geometry while preserving semantic fidelity. In particular, inter-class discrimination is enhanced through a concave distance transformation that enlarges pairwise margins between class-level embeddings. To improve intra-class compactness without collapsing meaningful semantic variation, we introduce a prototype-preserving formulation that identifies multiple semantic modes within each class and constrains their scatter around class centroids in the learned space. Together, these mechanisms produce a semantic representation that preserves meaningful class relationships while improving discriminative structure.

Once the \paperabb{} is constructed, frozen visual features extracted from image, video, or motion encoders are aligned to the learned semantic space through a lightweight MLP without fine-tuning the underlying backbone networks (Fig.~1b). This decouples semantic-space learning from visual feature extraction and enables improved classification performance while retaining the generalization capability of pretrained encoders.

We further provide theoretical analysis showing that the proposed framework preserves semantic consistency while enforcing enhanced inter-class separation and bounded intra-class variance in the learned semantic space.

Extensive experiments on skeleton-based human action recognition, video classification, and image classification benchmarks demonstrate the effectiveness and generality of \paperabb{} across multiple modalities and architectures. The proposed framework consistently improves classification performance over existing semantic alignment methods, achieving gains exceeding 3 percentage points on average and up to 8.5 points on challenging benchmarks.

The main contributions of this work are summarized as follows:
\begin{itemize}
    \item We propose \paperabb{}, a framework for learning a discriminative low-dimensional semantic space for visual classification that jointly enforces dimensional compatibility, semantic consistency, inter-class separation, and intra-class compactness.
    \item We introduce a geometry-transformation and prototype-preserving formulation that reshapes pretrained semantic embeddings into a task-adaptive discriminative semantic space.
    \item We provide theoretical analysis demonstrating that \paperabb{} preserves semantic structure while improving class separability and controlling intra-class variance.
\end{itemize}

\section{Related work}
\textbf{Dimensionality Reduction:} Existing methods often learn discriminative low-dimensional representations by increasing class separability and reducing class variance, but they do not preserve semantics, as shown in Table~\ref{tab:dr_methods}. For instance, Linear Discriminant Analysis (LDA) \cite{lda} does not preserve global geometry, which is important for semantic preservation. LFDA \cite{LFDA}, MFA \cite{MFA}, and Graph-Regularized LDA \cite{graphlda} prioritize local discrimination and often lose global structure. Supervised Isomap \cite{isomap} enhances between-class separation but provides limited compactness and geometry preservation. Conversely, MDS \cite{mds} preserves global pairwise geometry but does not improve separation or compactness, while UMAP \cite{umap} and t-SNE \cite{tsne} \textit{neither guarantee global geometry nor explicitly enhance class discrimination}.

\textbf{Metric and prototype-based learning:} Metric and prototype-based learning improves classification through compact intra-class and separated inter-class embeddings \cite{kulis2013metric}. Supervised contrastive learning \cite{supcon} pulls same-class samples together and pushes negative samples apart, while Multi-Similarity \cite{msloss} and Circle Loss \cite{circleloss} focus on hard pairs. Proxy Anchor Loss \cite{proxyanchor} learns class proxies, and ArcFace \cite{arcface} adds an angular margin for sharper boundaries. However, \textit{these methods improve discrimination without preserving semantics}.

\textbf{Text-guided visual classification.}
Text-guided classification aligns visual and textual features for semantic motion, video, and image recognition. In image classification, CLIP \cite{clip} aligns image-text features via contrastive learning, while CoOp \cite{coop}, CoCoOp \cite{cocoop}, MaPLe \cite{maple}, and PromptSRC \cite{promptsrc} adapt and regularize textual semantics for downstream tasks. Tip-Adapter \cite{tip} and WATT \cite{watt} further improve classification using multiple textual descriptions. For motion classification, ActionCLIP \cite{actionclip} extends CLIP to video recognition, while GAP \cite{gap} and CrossGLG \cite{crossglg} align skeleton features with textual semantics. In zero-shot recognition, PURLS \cite{purls}, Neuron \cite{neuron}, and FS-VAE \cite{FSVAE} leverage motion-phase or frequency-aware text alignment to distinguish similar actions. However, \textit{these methods mainly improve semantic alignment without explicitly enhancing target-space discrimination}, which limits class separability.

\section{Problem formulation} \label{sec:problem_formulation}
A semantic space is a high dimensional vector space in which linguistic (or other meaningful) items are represented as vectors, such that semantic relatedness is captured by geometric proximity (e.g. distance or angle) between those vectors \cite{Ethayarajh2019HowCA, karlgren2021semantics}.
Let $\mathcal{C}=\{1,\dots,C\}$ denote a set of semantic classes. A collection of semantic embeddings is denoted as $\{(e_i,c_i)\}_{i=1}^n$, where $c_i \in \mathcal{C}$ is its associated class, and $E=\{e_i\}_{i=1}^n$, $e_i \in \mathbb{R}^{d_{E}}$, denote the set of embeddings . 

Let \(\mathcal{E}\subseteq \mathbb{R}^{d_E}\) denote the original semantic space produced by a pretrained text encoder, where \(d_E\) is the semantic embedding dimension. Let \(\rho_E:\mathcal{C}\times\mathcal{C}\to\mathbb{R}_{\ge 0}\) denote the semantic dissimilarity metric induced by the original semantic embeddings. The objective is to learn a mapping, $g_{\theta}: \mathbb{R}^{d_E} \rightarrow \mathbb{R}^{d_v}$ that maps the original semantic embeddings into the \paperabb{}. 
Here, \(d_v\) denotes both the visual feature dimension and the \paperabb{} dimension, since the \paperabb{} is constructed to serve as the direct alignment target for visual features. The \paperabb{} is designed to satisfy three properties: dimensional compatibility with visual features, enhanced inter-class separation and intra-class compactness, and preservation of semantic fidelity.

\begin{definition}[Semantic Space with Semantic Fidelity]
\label{def:semantic-space}

Let \(\mathcal{P} = \{(c,c') : 1 \le c < c' \le C\}\) denote the set of all unordered class pairs. A target space \((\mathcal{C}, \rho_M)\) is said to maintain \emph{semantic fidelity} with respect to \((\mathcal{C}, \rho_E)\) if, for any \((c,c'), (k,k') \in \mathcal{P}\),
\begin{equation}
    \rho_E(c,c') < \rho_E(k,k')
    \implies
    \rho_M(c,c') < \rho_M(k,k').
\end{equation}
\end{definition}

This definition yields a task-agnostic criterion: the relative order of class dissimilarities induced by \(\rho_E\) is preserved under \(\rho_M\), even when the absolute geometry of the space changes.

\begin{condition}[Sufficient Condition for Semantic Fidelity] Let
\label{cond:semanticity}
\begin{equation}
    \Delta_{\min}
    =
    \min_{\substack{(c,c'),(k,k')\in\mathcal{P}\\ t_{cc'} \neq t_{kk'}}}
    \big| t_{cc'} - t_{kk'} \big|,
\end{equation}
denote the minimum separation between distinct classes in the target space. The worst case approximation error with respect to the target distances is defined as,
\begin{equation}
\varepsilon_\infty
=
\max_{(c,c') \in \mathcal{P}}
\big| \rho_M(c,c') - t_{cc'} \big|.
\end{equation}
If $\varepsilon_\infty < \frac{\Delta_{\min}}{2}$, then the derived semantic space \((\mathcal{C}, \rho_M)\) maintains semantic fidelity with respect to \((\mathcal{C}, \rho_E)\).
\end{condition}

\section{Learning a task-adaptive semantic space (\paperabb{})} 

\paragraph{Semantic geometry enhancement:}
To improve inter-class separation while preserving semantic fidelity, a target inter-class distance configuration is constructed to expand distances between class level representations. The idea is to apply an order preserving transformation to the distances of the semantic space, thereby increasing separation among nearby classes without altering their relative semantic ordering.

This can be achieved by applying an increasing concave transformation to the reference inter-class distances \(\delta_{cc'} = \rho_E(c,c')\) to obtain the target distances \(t_{cc'}\):
\begin{equation}
    t_{cc'} = f_a(\delta_{cc'})
            = \delta_{\max} \Big(\tfrac{\delta_{cc'}}{\delta_{\max}}\Big)^a,
\end{equation}
where $a \in (0,1]$ controls the degree of concavity and $\delta_{\max} = \max_{c<c'} \delta_{cc'}$. Since $f_a$ is strictly increasing, the relative ordering of inter-class distances is preserved, thereby maintaining semantic fidelity.

The transformation $f_a(x)=\delta_{\max}(x/\delta_{\max})^a$ is strictly increasing for $a\in(0,1]$, preserving the relative ordering of inter-class semantic dissimilarities and ensuring that the transformed distances ${t_{cc'}}$ satisfy Definition~\ref{def:semantic-space}.

In addition, For $a<1$, the ratio $(x/\delta_{\max})^{a-1}$ decreases with $x$, causing smaller distances to be amplified more than larger ones. This selectively increases separation among semantically similar classes while minimally affecting dissimilar pairs.

\paragraph{Learning the target semantic geometry:}
Let \(m_i = g_\theta(e_i)\in\mathbb{R}^{d_v}\) denote the mapped semantic embedding in the \paperabb{}. Class level representations are obtained by computing centroids over the embeddings belonging to each class, denoted by $\mu_c^M$. These class centroids lie on the unit hypersphere \(\mathbb{S}^{d_v-1}\) and induce the learned \paperabb{} metric \(\rho_M\) via angular distance: $
\rho_M(c,c')=\arccos(\mu_c^M\cdot\mu_{c'}^M)$

To align the learned class centroids with the target inter-class distance configuration $\{t_{cc'}\}$, a stress based objective is minimized:
\begin{equation}
    \mathcal L_{\mathrm{sem\_geo}}
    =
    \frac{1}{|\mathcal P|}
    \sum_{(c,c')\in\mathcal P}
    \left(
        \arccos(\mu_c^M \cdot \mu_{c'}^M)
        - t_{cc'}
    \right)^2.
\end{equation}
This objective penalizes deviations between the learned and target inter-class distances, driving the learned semantic metric to approximate the target distance configuration.

\paragraph{Compactness enhancement via multi-prototype regularization:}
While the separation objective enhances inter-class separation, it does not constrain within-class scatter, which can lead to overlap between nearby classes. 
To address this issue, we introduce a compactness regularization that controls intra-class geometry while preserving meaningful variation or style within each class.

We represent each class using \(K\) prototypes. Each prototype captures one mode of semantic variation, such as styles, within the class. The prototypes are obtained through clustering, such as \(k\)-means, to the original semantic embeddings of each class in \(\mathbb{R}^{d_E}\), and are then mapped into the \paperabb{}. 

To enforce compactness, two constraints are introduced. First, each embedding \(m_i=g_\theta(e_i)\) is forced to remain close to its assigned prototype $\pi_i$ as,

\begin{equation}
\mathcal{L}_{\mathrm{local}}
=
\frac{1}{n}
\sum_{i=1}^{n}
\left\|m_i - p_{c_i,\pi_i}^M\right\|_2^2,
\end{equation}
where \(c_i\) is the class label of embedding \(e_i\), and \(p_{c_i,\pi_i}^M\) denotes the \paperabb{} prototype assigned to \(m_i\).

Second, the prototype centers of each class are regularized to remain compact around the class centroid \(\mu_c^M\). Specifically, we constrain the centroid-to-prototype distances to remain smaller than the global classwise separation in the target space, thereby reducing overlap between classes. This is achieved by assigning each prototype a target centroid-to-prototype distance \(r_{c,j}\), chosen to keep prototypes compact relative to the minimum target inter-class separation, and penalizing deviations from it:
\begin{equation}
\mathcal{L}_{\mathrm{proto}}
=
\frac{1}{C}\sum_{c\in\mathcal{C}}\frac{1}{K}
\sum_{j=1}^{K}
\left(
\left\|p_{c,j}^M-\mu_c^M\right\|_2-r_{c,j}
\right)^2 .
\end{equation}
Together, a semantic variation or style is constructed, $\mathcal{L}_{\mathrm{sem\_sty}}=\mathcal{L}_{\mathrm{local}}+\mathcal{L}_{\mathrm{proto}}$

\paragraph{Total Objective \& Algorithm:}
The \paperabb{} mapping $g_\theta$ is trained with
$\mathcal{L}_{\mathrm{\paperabb{}}}=\mathcal{L}_{\mathrm{sem\_geo}}+\lambda \mathcal{L}_{\mathrm{sem\_sty}}$,
where $\lambda=1$ balances semantic geometry and style preservation. Algorithmic details are given in Appendix~\cref{alg}.

\section{Theoretical properties of the \paperabb{}}
\textbf{Proposition 1 (Semantic fidelity)}  When Condition~\ref{cond:semanticity} holds, the derived semantic space \((\mathcal{C}, \rho_M)\) reflects the relative ordering of inter-class semantic dissimilarities induced by the original semantic metric \(\rho_E\). \newline
\textbf{Proposition 2 (Selective separation enhancement)}  
For concavity parameter $a<1$, the target distance transformation expands smaller reference inter-class distances proportionally more than larger ones, thereby increasing separation among semantically similar classes. \newline
\textbf{Proposition 3 (Effect of compactness regularization)}
Under the assumption that multiple semantic embeddings associated with the same class capture stylistic variations, prototype-based compactness regularization acts on intra-class dispersion and introduces no explicit cross-class interactions.\newline
Formal statements and proofs are provided in Appendix \cref{app:theory} of the supplementary material.

\section{Application to visual classification}
\label{sec:har-application}

To construct a \paperabb{} for a visual task, $l$ textual descriptions are generated per class using a large language model to capture class specific variations. The descriptions are encoded using a pretrained text encoder such as CLIP \cite{clip}, BERT \cite{bert}, BGE \cite{bge}, and the resulting embeddings are processed using \paperabb{} Algorithm (\ref{alg:desed}). Prompting and text encoding details are provided in Appendix \cref{sec:text_generation}.
 
Let $f_\phi$ denote a frozen visual encoder that maps input visual content $x$ to a feature vector $v=f_\phi(x)\in\mathbb{R}^{d_v}$. The visual-to-\paperabb{} alignment function is denoted by \(h_\psi:\mathbb{R}^{d_v}\rightarrow\mathbb{R}^{d_v}\), which aligns visual features to the $K$ prototypes of the \paperabb{} via $h_\psi$, using an InfoNCE-style objective. Details of the alignment and inference are provided in Appendix \cref{sec:har-application}. 

\section{Experiments}
The proposed method has been validated on a variety of classification and datasets. They are \textbf{Skeleton-based human action recognition} (HAR) on \textit{NTU RGB+D} \cite{ntu60} (NTU-60), a 3D action dataset with 60 actions, performed with 40 different subjects; \textit{NTU RGB+D 120} \cite{ntu120} (NTU-120), an extension of the NTU-60 with additional 60 actions, with 106 different subjects; \textbf{Video classification} on \textit{HMDB51} \cite{hmdb} contains 6,766 video clips of 51 action categories, and on \textit{Aridv1.5} \cite{arid} contains 3,784 video clips of 11 action categories; \textbf{Image classification} on \textit{CIFAR-10} \cite{cifar10}, a widely used image classification dataset with 60,000 low-resolution RGB images across 10 classes, \textit{CUB200} \cite{cub200}, a widely used fine-grained image classification dataset with 200 bird species, and \textit{Oxford-IIIT Pet} \cite{oxfordiit} is a fine-grained image classification dataset of 37 pet categories including dog and cat breeds.

\paragraph{Implementation:} \label{sec:implement_det}
All experiments are implemented in PyTorch and conducted on a single NVIDIA RTX 4090 GPU. Class-level text descriptions are generated using GPT such as GPT-5, Gemini \cite{gemini}, Claude \cite{claude}, and encoded with a pretrained text encoders such as CLIP \cite{clip}, BERT \cite{bert} and BGE \cite{bge}. The prompt template used for text generation is provided in Appendix \cref{sec:text_generation}. For evaluation, pretrained weights from the original repositories were used for all encoders. Specifically, USDRL \cite{foundskel} and MAMP \cite{mamp} were used as motion encoders; VideoMAEv2 \cite{videomae} and SMILE \cite{thoker2025smile} with ViT-B/16 were used as video encoders; and DINOv3 \cite{dinov3} with ViT-S/16 and MAE \cite{mae} with ViT-B/16 were used as image encoders.

\begin{table}[t]
\centering
\caption{\small 
\paperabb{} alignment improves visual classification across HAR, video, and image classification tasks, datasets, encoder architectures. Each row evaluates a frozen encoder under four settings: linear probing, linear projection to \paperabb{}, MLP alignment to the original text-encoder semantic space, and MLP alignment to \paperabb{}. The final column reports the \paperabb{} gain over MLP alignment to the original text-encoder semantic space on the hardest 30\% of classes. Here SSL, SL, Trans., GCN, and Align. denote self-supervised learning, supervised learning, Transformer, graph convolutional network, and alignment, respectively.}
\label{tab:skeleton_encoders}
\setlength{\tabcolsep}{2pt}
\renewcommand{\arraystretch}{1.1}
\resizebox{\columnwidth}{!}{%
\begin{tabular}{l l l l l l c c c c c}
\toprule
\multirow{2}{*}{Task} 
& \multirow{2}{*}{Dataset} 
& \multirow{2}{*}{Encoder} 
& \multirow{2}{*}{Publication} 
& \multirow{2}{*}{Arch.} 
& \multirow{2}{*}{Pretrain} 
& \multicolumn{4}{c}{Top-1 Accuracy} 
& \multirow{2}{*}{\makecell{Hard-class\\Gain}} \\
\cmidrule(lr){7-10}
& & & & & 
& \makecell{Linear\\Probe} 
& \makecell{\paperabb{}\\Linear} 
& \makecell{Text-Enc.\\MLP Align.} 
& \makecell{\paperabb{}\\MLP Align.} 
& \\
\midrule

\multirow{6}{*}{HAR}
&    & USDRL  & PAMI'25 & Trans. & SSL & 84.2 & 85.2 & 85.0 & \textbf{85.6} & 2.6 \\
&  NTU60      & MAMP   & ICCV'23 & Trans. & SSL & 84.9 & 87.1 & 85.0 & \textbf{88.2} & 6.3 \\
&         & CTR-GCN & ICCV'21 & GCN   & SL  & 89.9 & 90.6 & 90.0 & \textbf{90.9} & 1.1 \\ 
\cline{2-11}
&   & USDRL  & PAMI'25 & Trans. & SSL & 74.9 & 76.5 & 75.4 & \textbf{76.7} & 4.0 \\
&  NTU120       & MAMP   & ICCV'23 & Trans. & SSL & 78.6 & 80.8 & 79.1 & \textbf{81.0} & 4.1 \\
&         & CTR-GCN & ICCV'21 & GCN   & SL  & 84.9 & 85.7 & 85.3 & \textbf{85.9} & 1.4 \\ 

\midrule

\multirow{4}{*}{Video}
& HMDB51   & SMILE      & CVPR'25 & Trans. & SSL & 70.8 & 71.5 & 71.5 & \textbf{71.8} & 0.9 \\
&          & VideoMAEv2 & CVPR'23 & Trans. & SSL & 83.2 & 84.8 & 84.0 & \textbf{84.9} & 0.9 \\ 
\cline{2-11}
& Aridv1.5 & SMILE      & CVPR'25 & Trans. & SSL & 71.5 & 76.7 & 74.2 & \textbf{76.9} & 1.5 \\
&          & VideoMAEv2 & CVPR'23 & Trans. & SSL & 79.0 & 79.9 & 79.3 & \textbf{80.1} & 0.8 \\ 

\midrule

\multirow{6}{*}{Image}
& CIFAR10     & DINOv3 & arXiv'25 & Trans. & SSL & 96.6 & 97.7 & 97.4 & \textbf{98.0} & 0.7 \\ 
&              & MAE    & CVPR'22  & Trans. & SSL & 84.2 & 86.0 & 84.8 & \textbf{89.5} & 6.5 \\ 
\cline{2-11}
& CUB-200      & DINOv3 & arXiv'25 & Trans. & SSL & 84.8 & 88.2 & 85.5 & \textbf{88.4} & 8.5 \\ 
&              & MAE    & CVPR'22  & Trans. & SSL & 30.7 & 38.5 & 35.6 & \textbf{40.2} & 8.3 \\ 
\cline{2-11}
& Oxford & DINOv3 & arXiv'25 & Trans. & SSL & 94.1 & 95.0 & 94.4 & \textbf{95.1} & 1.6 \\ 
& -IIIT Pet        & MAE    & CVPR'22  & Trans. & SSL & 80.9 & 82.8 & 82.8 & \textbf{84.3} & 1.3 \\ 

\bottomrule
\end{tabular}%
}
\end{table}

\paragraph{Results:} \label{sec:results}
The effectiveness of \paperabb{} alignment is evaluated across tasks, datasets, and encoder families, as summarized in \cref{tab:skeleton_encoders}. Four settings are compared: \textit{Linear probe}, a supervised linear classifier on frozen features without semantic alignment; \textit{\paperabb{} linear}, a linear projection to \paperabb{}; \textit{Text-Enc. MLP Align.}, MLP alignment to the original text-encoder semantic space; and \textit{\paperabb{} MLP Align.}, MLP alignment to \paperabb{}. 

\textbf{\textit{Comparison with the original text-encoder semantic space}:} 
To determine whether the gains come from semantic supervision alone or from constructing a discriminative target semantic space, we align frozen visual features using the same projection layer to either the original text-encoder semantic space (\textit{Text-Enc. MLP Align.}) or \paperabb{} (\textit{\paperabb{} MLP Align.}). \textit{\paperabb{} MLP Align.} consistently outperforms \textit{Text-Enc. MLP Align.} across all datasets and encoders; for example, it improves NTU120 accuracy with the frozen USDRL encoder by 1.3 percentage points. Notable gains occur for semantically similar action pairs, such as ``put object into bag''/``take object out of bag'' and ``thumbs up''/``thumbs down'', which are weakly separated in the original text-encoder space. By enhancing inter-class separation while preserving semantic fidelity, \paperabb{} provides a more discriminative alignment target.

For fine-grained image classification, \paperabb{} improves over \textit{Text-Enc. MLP Align.} by 2.9 percentage points on CUB200 with the frozen DINOv3 encoder. Large gains appear among visually and semantically similar bird groups, such as gulls (e.g., ``California gull'', ``Herring gull''), terns (e.g., ``Arctic tern'', ``Black tern''), and sparrows (e.g., ``Baird sparrow'', ``Brewer sparrow''). These species share common visual and semantic attributes and are therefore weakly separated in the original text-encoder semantic space; such fine-grained distinctions particularly benefit from the proposed discriminative alignment.

\textbf{\textit{Hard-class performance}:} Hard-class gains are reported for the hardest 30\% of classes, defined as those with the lowest per-class accuracy under baseline \textit{Text-Enc. MLP Align.}. As shown in \cref{tab:skeleton_encoders}, \textit{\paperabb{} MLP Align.} improves over \textit{Text-Enc. MLP Align.} on these hard classes, with maximum gains of 6.3, 1.5, and 8.5 percentage points on HAR, video, and image classification, respectively. These gains demonstrate that \paperabb{} provides a more discriminative semantic space for ambiguous classes.

\textbf{\textit{Comparison with metric-learning baselines}:} Since \paperabb{} aims to improve discriminative visual representations, it is compared with SupCon~\cite{supcon}, Multi-Similarity~\cite{msloss}, Circle loss~\cite{circleloss}, and ArcFace~\cite{arcface} under the same frozen-DINOv3 setting on CUB200. This comparison tests whether learning a discriminative embedding space from frozen visual features alone can match \paperabb{} alignment. Unlike \paperabb{}, these baselines use class-label supervision to improve discrimination without explicitly retaining class-wise semantics.

As shown in \cref{tab:metric_learning}, all metric-learning baselines improve over the linear probe, confirming the value of discriminative supervision. However, \paperabb{} achieves the best Top-1 accuracy, outperforming the strongest metric-learning baseline, Circle loss, by 1.7 percentage points. This suggests that the gain is not only from enforcing compactness and separation using visual labels, but also from aligning visual features to a semantic target space designed to improve discrimination.

\paragraph{Ablation Study:}
Ablation studies are conducted on NTU60 and CIFAR-10 using frozen MAMP and MAE encoders, respectively. Best results are shown in \textbf{bold}. \newline
\textbf{\textit{Impact of projection layer architecture:}}
\Cref{tab:projection_layer} compares linear and MLP-based projections for aligning frozen features to the \paperabb{}. A linear projection already improves over the baseline, indicating that the \paperabb{} provides a useful target space. The best results are obtained with a pre-normalized three-layer MLP, LN-MLP-3, using a Linear-GELU-Linear-GELU-Linear design, achieving 3.3 and 5.0 percentage-point gains over the baseline on HAR and image classification, respectively. This experiment suggests that moderate nonlinear alignment is beneficial, while deeper projection heads provide no additional gain.

\textbf{\textit{Impact of separation and compactness}:} \Cref{tab:proto_align} shows the step-wise effect of each operation. Alignment without separation provides general semantic-alignment reference and improves over the baseline by 0.1 and 0.4 percentage points on HAR and image classification. Adding separation with a single prototype further improves performance by 2.6 and 3.8 percentage points, showing the benefit of a more separated target space. The best results are obtained with multi-prototype \paperabb{}, reaching 88.2 and 89.2, demonstrating that modeling multiple intra-class modes together with separation produces the most discriminative target space.

\textbf{\textit{Impact of the number of prototypes}:} \Cref{tab:num_prototypes} reports the effect of varying the number of prototypes used to represent \paperabb{}. Changing the number of prototypes changes the number of modes used to model class-specific variations. In our implementation, the prototypes are obtained by varying the number of clusters in the \(k\)-means clustering. We evaluate prototype counts from 2 to 5, and observe the best classification accuracy when 4 prototypes are used.

\textbf{\textit{Impact of separation enhancement levels}:} \Cref{tab:alpha_ablation} reports classification performance under different levels of separation in the \paperabb{}. Here, the pairwise distances between classes are controlled through a concave transformation parameterized by $a$, where smaller values of $a$ increase inter-class separation and larger values reduce it. As $a$ decreases, the \paperabb{} gradually approaches a uniform geometric distribution. The best performance 88.2\%, 89.2\% is achieved for HAR and image classification respectively at $a = 0.2$. 

\textit{Additional ablation studies on the \paperabb{} learning, the number of text descriptions per class, the choice of text-generating LLM, and the choice of text encoder are provided in Appendix~\cref{app:ablation_st}.} 


\begin{table*}[t]
\centering
\setlength{\tabcolsep}{2.3pt}
\renewcommand{\arraystretch}{1.05}
\begin{minipage}[t]{0.24\textwidth}
\centering
\caption{\small Comparison with metric-learning methods on CUB-200 using DINOv3.}
\label{tab:metric_learning}
\begin{tabular}{lc}
\toprule
Method & Top-1 \\
\midrule
Linear & 84.4 \\
SupCon & 86.3 \\
Multi-Sim. & 86.4 \\
Circle & 86.7 \\
ArcFace & 86.2 \\ \midrule
\paperabb{} (ours) & \textbf{88.4} \\
\bottomrule
\end{tabular}
\end{minipage}
\begin{minipage}[t]{0.45\textwidth}
\centering
\caption{\small Projection design ablation. $h$, LN, and $d_v$ denote hidden dimension, layer normalization, and visual feature dimension.}
\label{tab:projection_layer}
\begin{tabular}{llcccc}
\toprule
Variant & Design & $h$ & LN & HAR & Img. \\
\midrule
Baseline & -- & -- & -- & 84.9 & 84.2 \\
\midrule
Linear & Linear & -- & No & 87.1 & 86.0 \\
\midrule
\multirow{5}{*}{MLP}
& MLP-2 & $d_v$ & No & 87.5 & 87.2 \\
& LN-MLP-2 & $d_v$ & Yes & 87.9 & 88.3 \\
& LN-MLP-2 & $d_v/2$ & Yes & 87.8 & 87.0 \\
& LN-MLP-3 & $d_v$ & Yes & \textbf{88.2} & \textbf{89.2} \\
& LN-MLP-4 & $d_v$ & Yes & 88.0 & 89.1 \\
\bottomrule
\end{tabular}

\vspace{1mm}
\begin{minipage}{0.96\linewidth}
\scriptsize
MLP-$n$ denotes $n$ linear layers with GELU between consecutive linear layers.
\end{minipage}
\end{minipage}
\hfill
\begin{minipage}[t]{0.28\textwidth}
\centering
\caption{\small Impact of separation and compactness.}
\label{tab:proto_align}
\setlength{\tabcolsep}{2pt}
\renewcommand{\arraystretch}{0.95}
\begin{tabular}{@{}>{\raggedright\arraybackslash}p{0.62\linewidth}cc@{}}
\toprule
Method & HAR & Img. \\
\midrule
Baseline & 84.9 & 84.2 \\
\midrule
+ alignment w/o separation & 85.0 & 84.8 \\
\midrule
+ single-prototype + separation & 87.6 & 88.6 \\
\midrule
+ multi-prototype + separation & \textbf{88.2} & \textbf{89.2} \\
\bottomrule
\end{tabular}
\end{minipage}
\end{table*}

\begin{table*}[t]
\setlength{\tabcolsep}{2.5pt}
\renewcommand{\arraystretch}{1.05}

\noindent\makebox[\textwidth][l]{%
\begin{minipage}[t]{0.25\textwidth}
\centering
\caption{\small Effect of the number of prototypes.}
\label{tab:num_prototypes}
\setlength{\tabcolsep}{3pt}
\begin{tabular}{>{\centering\arraybackslash}p{0.42\linewidth}cc}
\toprule
\multirow{2}{*}{\makecell{Prototype\\count}} & \multicolumn{2}{c}{Top-1} \\
\cmidrule(lr){2-3}
& HAR & Img. \\
\midrule
2 & 87.6 & 88.5 \\
3 & 87.8 & \textbf{89.2} \\
4 & \textbf{88.2} & \textbf{89.2} \\
5 & 88.1 & 89.0 \\
\bottomrule
\end{tabular}
\end{minipage}%
\hspace{0.02\textwidth}%
\begin{minipage}[t]{0.30\textwidth}
\centering
\caption{\small Impact of separation enhancement.}
\label{tab:alpha_ablation}
\setlength{\tabcolsep}{2pt}
\begin{tabular}{@{}lcc@{}}
\toprule
Method & HAR & Img. \\
\midrule
Baseline & 84.9 & 84.2 \\
Baseline + $a=1.0$ & 85.1 & 84.8 \\
Baseline + $a=0.3$ & 87.9 & 86.5 \\
Baseline + $a=0.2$ & \textbf{88.2} & \textbf{89.2} \\
Baseline + $a=0.1$ & 87.8 & 88.4 \\
\bottomrule
\end{tabular}
\end{minipage}%
\hspace{0.02\textwidth}%
\begin{minipage}[t]{0.38\textwidth}
\centering
\caption{\small Semantic fidelity evaluation on datasets with high semantic overlap.}
\label{tab:semantic_fidelity}
\setlength{\tabcolsep}{2pt}
\begin{tabular}{@{}lccc@{}}
\toprule
Metric & $k$ & NTU120 & CUB200 \\
\midrule
Spearman $\rho$ $\uparrow$ & -- & 0.99 & 1.00 \\
\midrule
\multirow{3}{*}{\makecell[l]{kNN-\\Overlap@k $\uparrow$}}
& 5   & 0.99 & 1.00 \\
& 10  & 0.99 & 1.00 \\
& 100 & 0.99 & 1.00 \\
\midrule
OVR $\downarrow$ & -- & 0.04\% & 0.00 \\
\bottomrule
\end{tabular}
\end{minipage}%
}
\end{table*}

\begin{table}[t]
\centering
\caption{T-to-M and T-to-I retrieval comparison: text encoder semantic-space vs \paperabb{} alignment}
\label{tab:retrieval_generalization}
\resizebox{\textwidth}{!}{
\begin{tabular}{lcccccc|lcccccc}
\toprule
\multicolumn{7}{c|}{T-to-M} & \multicolumn{7}{c}{T-to-I} \\
\cmidrule(lr){1-7} \cmidrule(lr){8-14}
\multirow{2}{*}{Alignment} 
& \multicolumn{3}{c}{$Precision@top\text{-}k\%$} 
& \multicolumn{3}{c|}{$mAP@top\text{-}k\%$}
& \multirow{2}{*}{Alignment} 
& \multicolumn{3}{c}{$Precision@top\text{-}k\%$} 
& \multicolumn{3}{c}{$mAP@top\text{-}k\%$} \\
\cmidrule(lr){2-4} \cmidrule(lr){5-7}
\cmidrule(lr){9-11} \cmidrule(lr){12-14}
& 1\% & 10\% & 50\% & 1\% & 10\% & 50\%
& & 1\% & 10\% & 50\% & 1\% & 10\% & 50\% \\
\midrule
text encoder & 0.97 & 0.95 & 0.87 & \textbf{0.97} & 0.93 & 0.84
& text encoder & 0.96 & 0.95 & 0.88 & 0.96 & 0.94 & 0.85 \\
\paperabb{}  & \textbf{0.98} & \textbf{0.97} & \textbf{0.92} & \textbf{0.97} & \textbf{0.95} & \textbf{0.90}
& \paperabb{}  & \textbf{0.98} & \textbf{0.97} & \textbf{0.94} & \textbf{0.97} & \textbf{0.96} & \textbf{0.93} \\
\bottomrule
\end{tabular}}
\end{table}


\paragraph{Semantic fidelity Validation experiments:}
Semantic fidelity between the original text-encoder space $\rho_E$ and the \paperabb{} space $\rho_M$ is evaluated using task-agnostic metrics on NTU120 and CUB200. Since the theoretical guarantee relies on a sufficient condition involving the maximum distance approximation error and the minimum target-distance gap, empirical validation is used to assess semantic fidelity in practical fine-grained settings. Following common evaluation practice, three complementary metrics are used: Spearman’s $\rho$ for preserving relative class-distance orderings across all class pairs~\cite{kendal,spearman}, k-nearest-neighbor (kNN) overlap for local geometric consistency between neighborhoods in $\rho_E$ and $\rho_M$~\cite{knn}, and ordinal violation rate (OVR) for the fraction of $\rho_E$-induced class-pair orderings reversed in $\rho_M$~\cite{ovr}. Formal definitions are provided in Appendix~\cref{app:semantic-validation}, and results are summarized in \cref{tab:semantic_fidelity}. On NTU120, high Spearman correlation and kNN overlap, together with a low OVR of 0.04\%, indicate strong semantic-fidelity preservation with only minor ordinal deviations. On CUB200, all metrics reach their maximum values with zero violations, indicating that \paperabb{} preserves the semantic fidelity of the text-encoder space.

\begin{figure*}[t]
    \centering

    \begin{subfigure}[b]{0.24\textwidth}
        \centering
        \includegraphics[width=\linewidth]{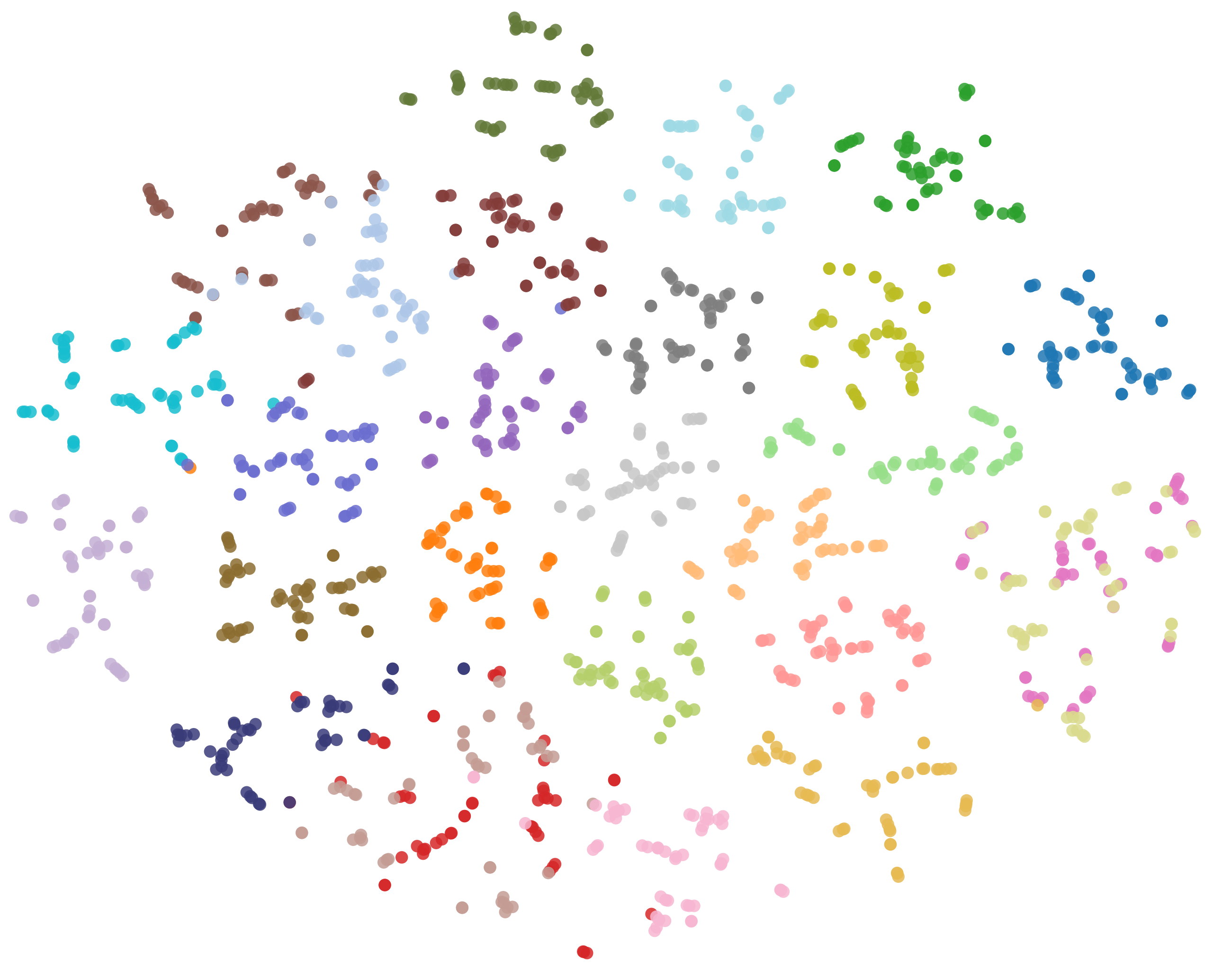}
        \caption{Original text space}
        \label{fig:sub1}
    \end{subfigure}
    \hfill
    \begin{subfigure}[b]{0.24\textwidth}
        \centering
        \includegraphics[width=\linewidth]{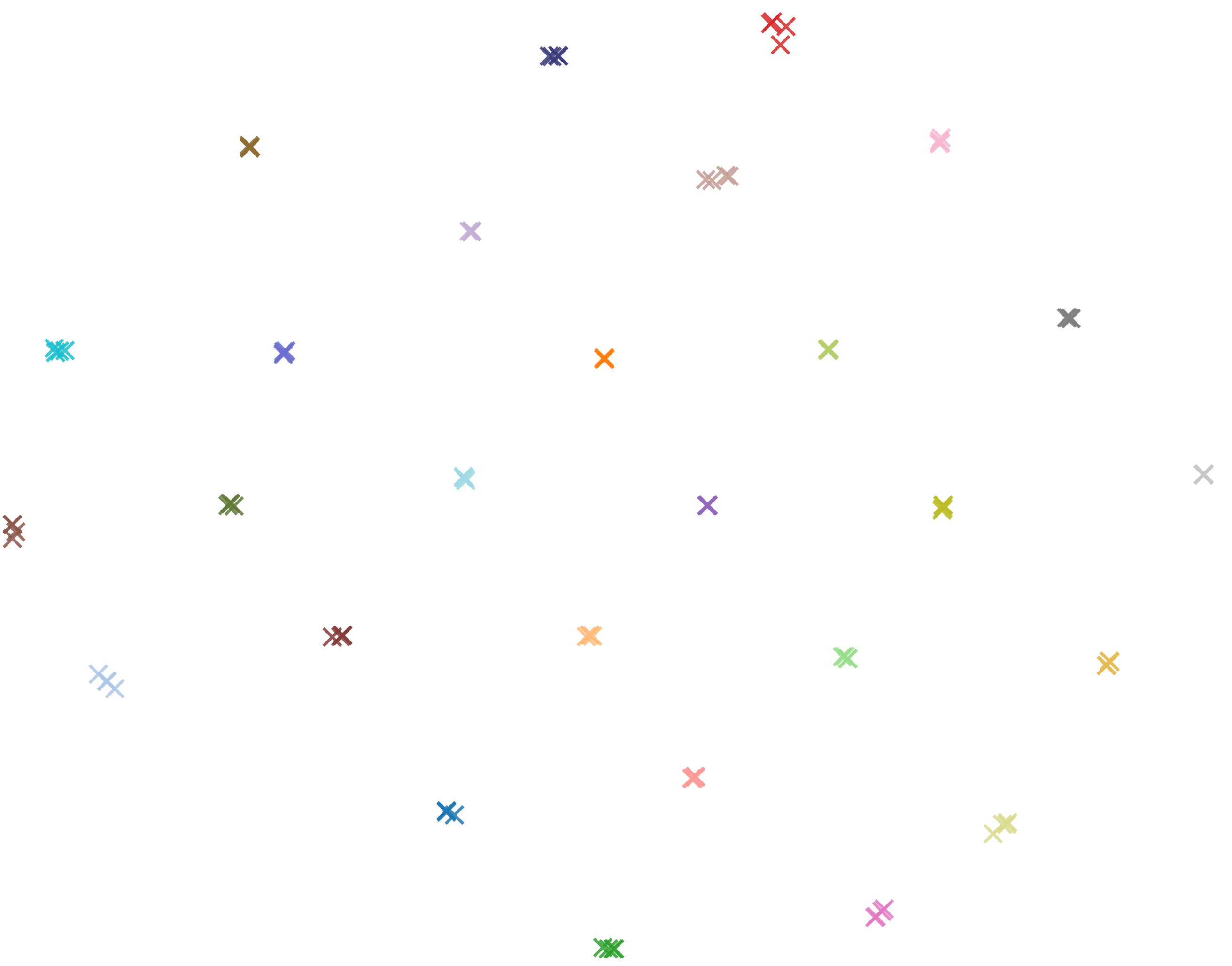}
        \caption{\paperabb{} space}
        \label{fig:sub2}
    \end{subfigure}
    \hfill
    \begin{subfigure}[b]{0.24\textwidth}
        \centering
        \includegraphics[width=\linewidth]{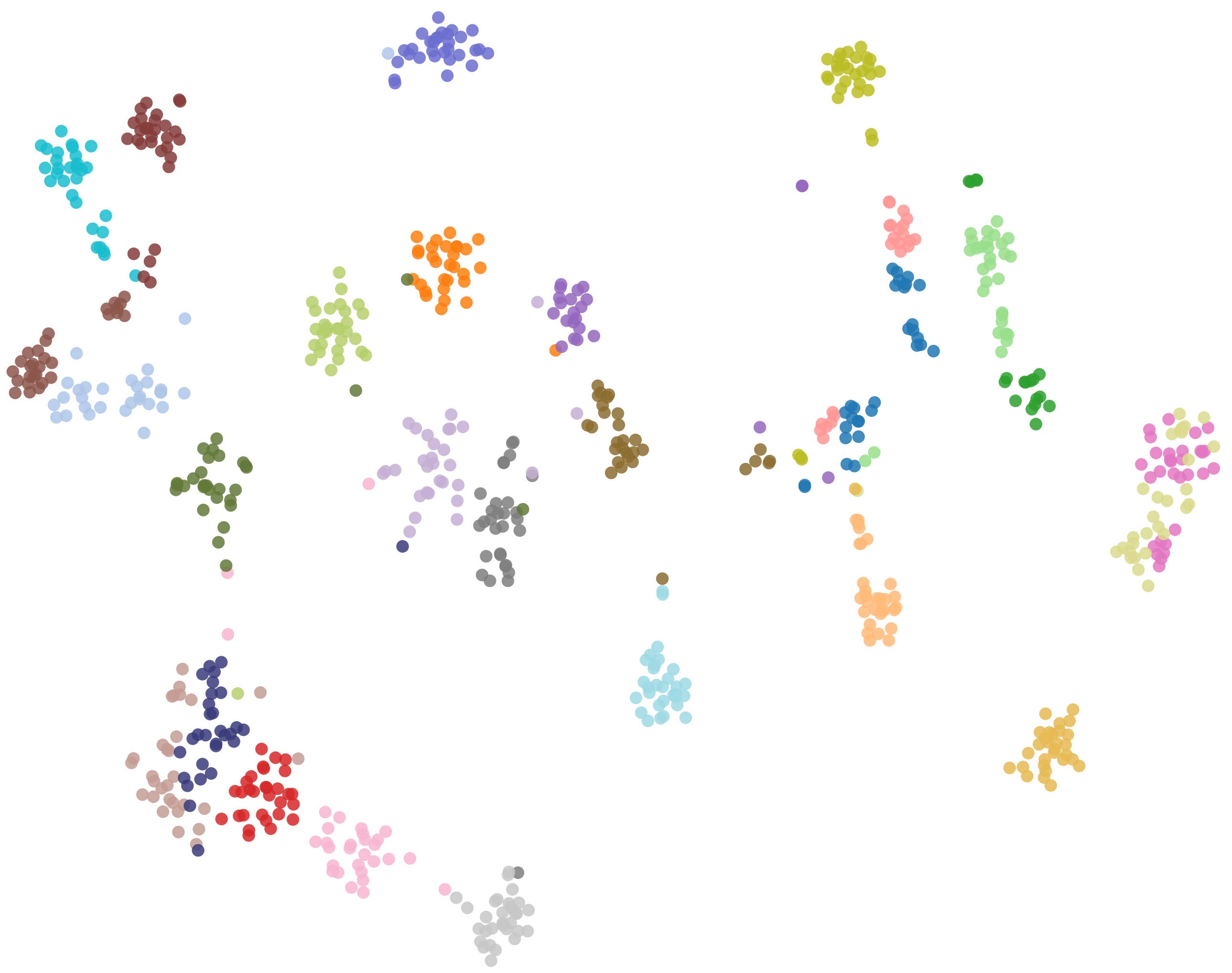}
        \caption{Baseline image space}
        \label{fig:sub3}
    \end{subfigure}
    \hfill
    \begin{subfigure}[b]{0.26\textwidth}
        \centering
        \includegraphics[width=\linewidth]{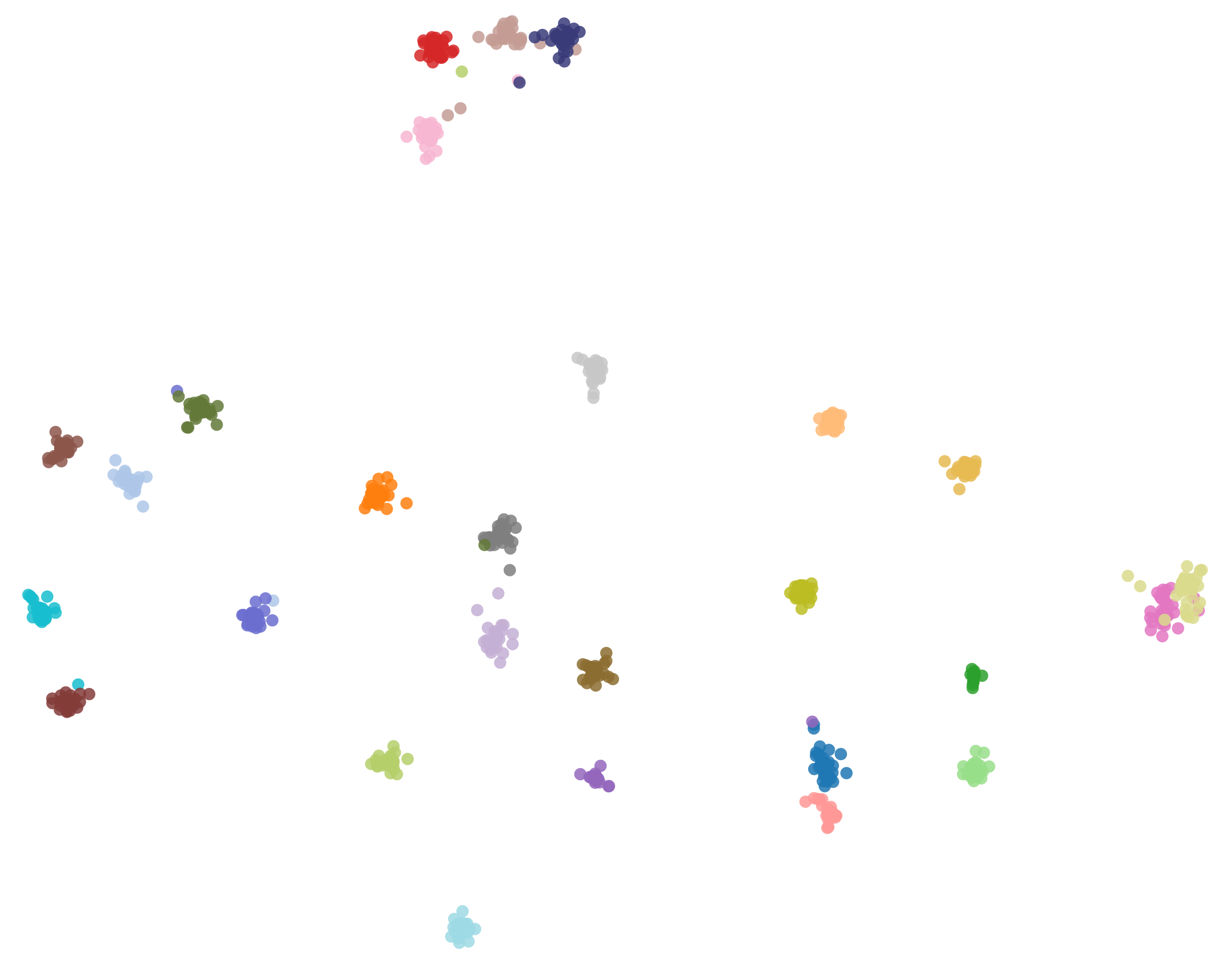}
        \caption{\paperabb{} aligned image space}
        \label{fig:sub4}
    \end{subfigure}

    \caption{
    t-SNE visualizations of semantic and image embeddings for 30 hard CUB200 classes. a) text-encoder semantic space shows scattered and partially overlapping clusters. b) The constructed \paperabb{} space exhibits improved separation and compactness while retaining semantic fidelity. c) Baseline frozen image embeddings show scattered and class overlaps. d) Image embeddings aligned to \paperabb{} form tighter and better-separated clusters.    }
    \label{fig:tsne_spaces}
\end{figure*}

\paragraph{Generalization evaluation:} \label{sec:generalisation}
We further evaluate \paperabb{} on text-to-motion (T-to-M) and text-to-image (T-to-I) retrieval to assess generalization beyond classification. Class descriptions are used as queries, and retrieval is performed using cosine similarity in the aligned space. USDRL and DINOv3 serve as frozen visual encoders for T-to-M on NTU120 and T-to-I on CUB200, respectively. Performance is measured using $Precision@top\text{-}k\%$ and $mAP@top\text{-}k\%$~\cite{eval_metric}.

As shown in \cref{tab:retrieval_generalization}, \paperabb{} consistently outperforms the original text-encoder semantic space across both retrieval tasks, with the largest gains at broader retrieval ranges. These results indicate that \paperabb{} preserves semantic structure while improving generalization beyond classification.

\paragraph{Visualization:}
\Cref{fig:tsne_spaces} shows t-SNE visualizations for the 30 hardest CUB200 classes selected using baseline DINOv3 accuracy. Compared with the original text-encoder semantic space (\cref{fig:sub1}), the constructed \paperabb{} space (\cref{fig:sub2}) exhibits improved inter-class separation and intra-class compactness while preserving semantic structure. Likewise, compared with frozen DINOv3 embeddings (\cref{fig:sub3}), \paperabb{}-aligned embeddings (\cref{fig:sub4}) form tighter and better-separated clusters, demonstrating the discriminative effectiveness of \paperabb{} for image classification.

\section{Conclusion, Limitations, and Future Work}
\label{sec:limitations}

We presented \paperabb{}, a framework for learning a discriminative low-dimensional semantic space for visual classification. \paperabb{} preserves semantic structure while improving inter-class separation and intra-class compactness through a geometry-aware transformation of pretrained text embeddings. Frozen visual features are aligned to \paperabb{} using a lightweight MLP, enabling consistent performance gains across skeleton-based action recognition, video classification, and image classification without fine-tuning pretrained encoders.

The quality of \paperabb{} depends on the coverage and reliability of the underlying text embeddings. In addition, the proposed semantic-fidelity guarantee preserves relative semantic relationships rather than the exact geometry of the original semantic space. The current framework also assumes a fixed class set, motivating future work on incremental and open-vocabulary settings. Extending \paperabb{} to adaptive higher-dimensional target spaces is another promising direction.

\appendix

\newpage

\setcounter{condition}{0}
\section*{Appendix}

\section{Additional Theoretical Analysis}
\label{app:theory}
This section provides supplementary derivations supporting the semantic fidelity
properties discussed in the main paper.

\subsection{Preliminaries and Notation}
We recall the key objects introduced in the main text.

\begin{itemize}
    \item $\mathcal{C} = \{1,\dots,C\}$: set of semantic classes.
    \item $\mathcal{P} = \{(c,c') : 1 \le c < c' \le C\}$: set of unordered class pairs.
    \item $\mu_c^E \in \mathbb{S}^{d_{E}-1}$: reference semantic centroid of class $c$.
    \item $\rho_E(c,c') = \delta_{cc'} = \arccos(\mu_c^E \cdot \mu_{c'}^E)$: semantic dissimilarity of class $c, c'$ in the original text-encoder semantic space.
    \item $\delta_{\max} = \max_{(c,c')\in\mathcal{P}} \delta_{cc'}$: maximum distance of class pairs in reference space.
    \item $t_{cc'} = f_a(\delta_{cc'})$ with
    $f_a(x) = \delta_{\max}(x/\delta_{\max})^a$, $a \in (0,1]$:
    target inter-class distances.
    \item $\mu_c^M \in \mathbb{S}^{d_v-1}$: learned class centroid in the derived semantic space (\paperabb{}).
    \item $\rho_M(c,c') = \arccos(\mu_c^M \cdot \mu_{c'}^M)$: learned \paperabb{} metric.
\end{itemize}

We further recall the quantities used to characterize approximate alignment between learned and target distances:
\begin{equation}
    \Delta_{\min}
    =
    \min_{\substack{(c,c'),(k,k')\in\mathcal{P}\\ t_{cc'} \neq t_{kk'}}}
    \big| t_{cc'} - t_{kk'} \big|,
\end{equation}
which denotes the smallest nonzero gap among target distances, and
\begin{equation}
\varepsilon_\infty
=
\max_{(c,c')\in\mathcal{P}}
\big| \rho_M(c,c') - t_{cc'} \big|,
\end{equation}
which bounds the worst-case deviation of learned distances from their targets.

\paragraph{Semantic fidelity.}
As defined in the main paper (Definition~\ref{def:semantic-space}), a derived semantic space \((\mathcal{C}, \rho_M)\) is said to maintain \emph{semantic fidelity} with respect to the reference space
\((\mathcal{C}, \rho_E)\) if, for all $(c,c'),(k,k')\in\mathcal{P}$,
\begin{equation}
    \rho_E(c,c') < \rho_E(k,k')
    \;\Rightarrow\;
    \rho_M(c,c') < \rho_M(k,k').
\end{equation}

\subsection{Derivation of the Semantic Fidelity Condition}
\label{app:semanticity-condition}

We restate Condition~\ref{cond:semanticity} from the main text and provide a formal derivation establishing its sufficiency for semantic fidelity

\begin{condition}[Semantic Fidelity Condition (restated)]
If
\begin{equation}
    \varepsilon_\infty < \frac{\Delta_{\min}}{2},
\end{equation}
then the learned \paperabb{} metric \(\rho_M\) does not invert any ordering of inter-class dissimilarities induced by the reference metric \(\rho_E\).
\end{condition}

\begin{lemma}[Order preservation under bounded deviation]
\label{lem:order-preservation}
Suppose that for all $(c,c')\in\mathcal{P}$,
\begin{equation}
    |\rho_M(c,c') - t_{cc'}| \le \varepsilon_\infty,
\end{equation}
and $\varepsilon_\infty < \Delta_{\min}/2$. Then, for any two pairs
$(c,c'),(k,k') \in \mathcal{P}$ satisfying
\begin{equation}
    t_{cc'} < t_{kk'},
\end{equation}
it holds that
\begin{equation}
    \rho_M(c,c') < \rho_M(k,k').
\end{equation}
\end{lemma}

\begin{proof}
Let $(c,c'),(k,k') \in \mathcal{P}$ with $t_{cc'} < t_{kk'}$. By definition of
$\Delta_{\min}$,
\begin{equation}
    t_{kk'} - t_{cc'} \ge \Delta_{\min}.
\end{equation}
Using the triangle inequality,
\begin{align}
    \rho_M(k,k') - \rho_M(c,c')
    &\ge (t_{kk'} - \varepsilon_\infty) - (t_{cc'} + \varepsilon_\infty) \\
    &= t_{kk'} - t_{cc'} - 2\varepsilon_\infty \\
    &\ge \Delta_{\min} - 2\varepsilon_\infty.
\end{align}
Since $\varepsilon_\infty < \Delta_{\min}/2$, the right-hand side is strictly positive, implying $\rho_M(c,c') < \rho_M(k,k')$.
\end{proof}

\begin{lemma}[Order preservation of target distances]
\label{lem:target-order}
Let $t_{cc'} = f_a(\rho_E(c,c'))$ with $f_a$ strictly increasing. Then, for any
$(c,c'),(k,k') \in \mathcal{P}$,
\begin{equation}
    \rho_E(c,c') < \rho_E(k,k')
    \;\Rightarrow\;
    t_{cc'} < t_{kk'}.
\end{equation}
\end{lemma}

\begin{proof}
Since $f_a$ is strictly increasing, it preserves the ordering of its inputs. Applying $f_a$ to both sides of $\rho_E(c,c') < \rho_E(k,k')$ yields $t_{cc'} < t_{kk'}$.
\end{proof}

\begin{proposition}[Semantic fidelity under bounded deviation]
\label{prop:semantic-fidelity}
If $t_{cc'}$ is constructed via a strictly increasing transformation of $\rho_E(c,c')$ and $\varepsilon_\infty < \Delta_{\min}/2$, then the derived semantic space $(\mathcal{C}, \rho_M)$ maintains semantic fidelity with respect to $(\mathcal{C}, \rho_E)$ in the sense of Definition~\ref{def:semantic-space}.
\end{proposition}

\begin{proof}
For any $(c,c'),(k,k') \in \mathcal{P}$ satisfying $\rho_E(c,c') < \rho_E(k,k')$, Lemma~\ref{lem:target-order} implies $t_{cc'} < t_{kk'}$. By Lemma~\ref{lem:order-preservation}, this ordering is preserved by $\rho_M$ under the stated bound on $\varepsilon_\infty$, yielding $\rho_M(c,c') < \rho_M(k,k')$. This establishes semantic fidelity.
\end{proof}
\subsection{Proofs of Theoretical Properties}
\label{app:proofs}
We now provide proofs of the theoretical properties stated in the main paper.

\textbf{Proposition 1 (Semantic fidelity).} \textit{If Condition~\ref{cond:semanticity} holds, then the derived semantic space $(\mathcal{C}, \rho_M)$ maintains semantic fidelity with respect to $(\mathcal{C}, \rho_E)$.}
\begin{proof}
Condition~\ref{cond:semanticity} asserts that the maximum deviation between the
learned inter-class distances $\rho_M(c,c')$ and their targets $t_{cc'}$ is bounded
by $\varepsilon_\infty < \Delta_{\min}/2$. By
Proposition~\ref{prop:semantic-fidelity} in
Appendix~\ref{app:semanticity-condition}, this bound guarantees that no ordering
of target distances is inverted by $\rho_M$. Since the target distances
$\{t_{cc'}\}$ preserve the ordering induced by $\rho_E$ via a strictly increasing
transformation, it follows that $(\mathcal{C}, \rho_M)$ maintains semantic fidelity
with respect to $(\mathcal{C}, \rho_E)$.
\end{proof}

\textbf{Proposition 2 (Selective separation enhancement).} \textit{For concavity parameter $a < 1$, the transformation $t_{cc'} = \delta_{\max}(\delta_{cc'}/\delta_{\max})^a$ expands smaller reference inter-class distances proportionally more than larger ones.}

\begin{proof}
Consider the ratio
\[
\frac{t_{cc'}}{\delta_{cc'}}
=
\left(\frac{\delta_{cc'}}{\delta_{\max}}\right)^{a-1}
=
\delta_{\max}^{\,1-a}\,\delta_{cc'}^{\,a-1}.
\]
For $a<1$, we have $a-1<0$. Since $\delta_{cc'}>0$, the derivative satisfies
\[
\frac{d}{d\,\delta_{cc'}}\!\left(\delta_{\max}^{\,1-a}\,\delta_{cc'}^{\,a-1}\right)
=
\delta_{\max}^{\,1-a}(a-1)\delta_{cc'}^{\,a-2}
<0,
\]
so the ratio is strictly decreasing in $\delta_{cc'}$. Hence, smaller reference distances receive a larger multiplicative expansion than larger ones, which selectively enhances separation among semantically similar classes.
\end{proof}

\textbf{Proposition 3 (Effect of prototype-based compactness regularization)}
\textit{Prototype-based compactness regularization operates only on intra-class dispersion and does not introduce explicit cross-class interactions.}

\begin{proof}
The term \(\mathcal{L}_{\mathrm{local}}\) depends only on each embedding \(m_i\) and its assigned prototype \(p_{c_i,\pi_i}^M\), while \(\mathcal{L}_{\mathrm{proto}}\) depends only on prototypes \(p_{c,j}^M\) and the centroid \(\mu_c^M\) of the same class. Hence, both terms act only within each class. Therefore, prototype-based compactness regularization reduces intra-class dispersion and introduces no explicit cross-class interactions.
\end{proof}

\section{Algorithm} \label{alg}
Algorithm~\ref{alg:desed} summarizes the \paperabb{} algorithm for learning the desired semantic space. The method first constructs a target inter-class distance configuration by applying a concave transformation to the distances in the semantic space. A mapping $g_\theta$ is then learned to align class-level representations with this target configuration while regularizing within class scatter. The resulting output is the \paperabb{}.

\begin{algorithm}[h]
\caption{\paperabb{}}
\label{alg:desed}
\begin{algorithmic}[1]
\Require Text embeddings $E=\{e_i\}_{i=1}^n$, labels $\{c_i\}$, concavity $a\in(0,1]$, number of prototypes $K$, steps $T$
\Statex
\State Define $\mathcal{I}_c \gets \{i \mid c_i=c\}$ for each class $c\in\mathcal{C}$
\State \textbf{Reference class centroids:} compute $\{\mu_c^E\}_{c=1}^C$ from $\{e_i\}$ and $\{\mathcal{I}_c\}$
\State \textbf{Reference distances:} $\delta_{cc'} \gets \arccos(\mu_c^E \cdot \mu_{c'}^E)$ for all $(c,c')\in\mathcal{P}$
\State $\delta_{\max} \gets \max_{(c,c')\in\mathcal{P}} \delta_{cc'}$
\State \textbf{Target distances:} $t_{cc'} \gets \delta_{\max}\big(\delta_{cc'}/\delta_{\max}\big)^a$ for all $(c,c')\in\mathcal{P}$
\State \textbf{Prototype initialization:} for each class \(c\), apply \(k\)-means to the original semantic embeddings to obtain \(K\) clusters and prototype assignments \(\{\pi_i\}\)
\State Initialize $g_\theta:\mathbb{R}^{d_{E}}\rightarrow\mathbb{R}^{d_v}$
\For{$t=1,\ldots,T$}
    \State \textbf{Project:} $m_i \gets g_\theta(e_i)$ for all $i$
    \State \textbf{Class centroids:} compute $\{\mu_c^M\}_{c=1}^C$ from $\{m_i\}$ and $\{\mathcal{I}_c\}$
    \State \textbf{Prototype centers:} compute $\{p_{c,j}^M\}$ from the mapped embeddings $\{m_i\}$ using the fixed prototype assignments $\{\pi_i\}$
    \Statex
    \State \textbf{Separation loss:}
     $\mathcal{L}_{\mathrm{sem\_geo}} \gets \frac{1}{|\mathcal{P}|}\sum_{(c,c')\in\mathcal{P}}
    \Big(\arccos(\mu_c^M\cdot\mu_{c'}^M) - t_{cc'}\Big)^2$
    \Statex
    \State \textbf{Local compactness loss:}
    $\mathcal{L}_{\mathrm{local}} \gets \frac{1}{n}\sum_{i=1}^{n}
\left\|m_i - p_{c_i,\pi_i}^M\right\|_2^2$
    \Statex
    \State \textbf{Prototype compactness loss:}
    $\mathcal{L}_{\mathrm{proto}} \gets \frac{1}{C}\sum_{c=1}^{C}\frac{1}{K}\sum_{j=1}^{K}
    \left(\left\|p_{c,j}^M-\mu_c^M\right\|_2-r_{c,j}\right)^2$
    \Statex
    \State \textbf{Update:} $\theta \gets \theta - \eta \nabla_\theta\big(\mathcal{L}_{\mathrm{sem\_geo}} + \lambda (\mathcal{L}_{\mathrm{local}} + \mathcal{L}_{\mathrm{proto}})\big)$
\EndFor
\State \Return \paperabb{} prototypes $\mathcal{P}^M=\{p^M_{c,j}\}_{c=1,j=1}^{C,K}$ 
\end{algorithmic}
\end{algorithm}

\section{Application to visual classification}
\label{sec:har-application}

To construct a \paperabb{} for a visual task, $l$ textual descriptions are generated per class using a large language model to capture class specific variations. The descriptions are encoded using a pretrained text encoder such as CLIP \cite{clip}, BERT \cite{bert}, BGE \cite{bge}, and the resulting embeddings are processed using \paperabb{} Algorithm \ref{alg:desed}. 
 
Let $f_\phi$ denote a frozen visual encoder that maps input visual content $x$ to a feature vector $v=f_\phi(x)\in\mathbb{R}^{d_v}$. The visual-to-\paperabb{} alignment function is denoted by \(h_\psi:\mathbb{R}^{d_v}\rightarrow\mathbb{R}^{d_v}\), which aligns visual features to \paperabb{} via $h_\psi$, which is typically an MLP. The \paperabb{} is constructed with the same dimensionality as the visual encoder output so that visual features and \paperabb{} prototypes lie in the same space \(\mathbb{R}^{d_v}\).

\paragraph{Multi-prototype alignment:}
Instance-level visual features generated by $f_\phi$ are aligned with $K$ prototypes of each class in the \paperabb{} using a learnable mapping function. Let
\(
P_c=\{p_{c,1},\dots,p_{c,K}\}\subset\mathbb{S}^{d_v-1}
\)
denote the prototypes associated with class \(c\) in the \paperabb{}. For notational simplicity, we omit the superscript \(M\) in this section and denote \paperabb{} prototypes by \(p_{c,j}\). Given a visual feature \(v\), the alignment function maps it to the \paperabb{} as $z = h_\psi(v), \: z \in \mathbb{S}^{d_v-1}$.

For a visual embedding \(z\) with ground-truth class \(c\), the positive prototype \(p^{+}\) is selected as the most similar prototype from the corresponding class, i.e. $p^{+}(z,c) = \arg\max_{p \in P_c} \; z^\top p.$

All prototypes belonging to the remaining classes are considered negatives. Based on this, the probability of assigning \(z\) to its positive prototype is defined using an InfoNCE-style objective:
\begin{equation}
    p\!\left(p^{+}\mid z\right)
    =
    \frac{\exp\!\left(z^\top p^{+}/\tau\right)}
    {\exp\!\left(z^\top p^{+}/\tau\right)
    +
    \sum_{c' \neq c}\sum_{j=1}^{K}
    \exp\!\left(z^\top p_{c',j}/\tau\right)},
\end{equation}
where \(\tau>0\) is a temperature parameter. The visual-semantic alignment loss is then given by 
\(
    \mathcal{L}_{\mathrm{vis}}
    =
    -\frac{1}{N}
    \sum_{i=1}^{N}
    \log p\!\left(p_i^{+} \mid z_i\right),
\)
where \(N\) is the number of samples.

\paragraph{Inference:} At inference, classification is performed by nearest-prototype matching based on maximum semantic similarity:
\(
    \hat{y}
    =
    \arg\max_{c\in\mathcal{C}} \max_{1\le j\le K} \; z^\top p_{c,j},
\)
where \(z=h_\psi(f_\phi(x))\) is the visual embedding aligned to the \paperabb{}, and \(p_{c,j}\) denotes the \(j\)-th \paperabb{} prototype of class \(c\).

\section{Semantic Validation Experiments}
\label{app:semantic-validation}
This appendix provides additional details on the evaluation metrics used to empirically validate semantic fidelity, as described in the main paper.

\paragraph{(1) Rank correlation with reference geometry:} Spearman’s rank correlation is computed between the rankings of inter-class distances induced by the reference semantic metric $\rho_E$ and those induced by the derived semantic metric $\rho_M$ over all class pairs.

\begin{itemize}
    \item \textbf{Spearman rank correlation.}
    Let $r_E(c,c')$ and $r_M(c,c')$ denote the ranks of $\delta_{cc'}$ and
    $\rho_M(c,c')$ among all pairs in $\mathcal{P}$. Spearman’s $\rho$ is defined as
    \begin{equation}
        \rho
        =
        1 - \frac{6}{|\mathcal{P}|(|\mathcal{P}|^2 - 1)}
        \sum_{(c,c')\in\mathcal{P}}
        \big(r_E(c,c') - r_M(c,c')\big)^2.
    \end{equation}
\end{itemize}

Higher values of $\rho$ indicate greater similarity between the rankings of inter-class distances induced by $\rho_E$ and $\rho_M$, providing an empirical proxy for semantic fidelity.

\paragraph{(2) kNN-Overlap@k:} kNN overlap evaluates, for each class $c\in\mathcal{C}$, the overlap between the sets of its $k$ nearest neighboring classes under the reference semantic metric $\rho_E$ and under the derived semantic metric $\rho_M$.

For each class $c\in\mathcal{C}$, let $\mathcal{N}^{E}_{K}(c)$ denote the set of the $K$ nearest classes to $c$ under the reference semantic metric $\rho_E$. Analogously, let $\mathcal{N}^{M}_{K}(c)$ denote the set of the $K$ nearest classes to $c$ under the derived semantic metric $\rho_M$.

\begin{equation}
\mathrm{kNN\text{-}Overlap@}K
:=
\frac{1}{C}
\sum_{c=1}^{C}
\frac{\left|\mathcal{N}^{E}_{k}(c)\cap \mathcal{N}^{M}_{K}(c)\right|}{K}.
\end{equation}

As a local comparison of nearest-neighbor relations, kNN overlap@K provides an empirical indicator of how local class neighborhoods induced by $\rho_E$ are reflected under $\rho_M$.

\paragraph{(3) Ordinal Violation Rate (OVR):}
The ordinal violation rate quantifies the fraction of ordered class-pair relations induced by $\rho_E$ that are violated under $\rho_M$. Let,

\begin{align}
N &:= \left|\left\{((c,c'),(k,k'))\in\mathcal{P}^2 : \rho_E(c,c') < \rho_E(k,k')\right\}\right|,\\
V &:= \left|\left\{((c,c'),(k,k'))\in\mathcal{P}^2 :
\rho_E(c,c') < \rho_E(k,k') \ \wedge\ \rho_M(c,c') \ge \rho_M(k,k')\right\}\right|.
\end{align}

Then $\mathrm{OVR} := V/N$.

OVR provides a violation-focused measure of ordinal relations induced by $\rho_E$ that are violated under $\rho_M$.

\section{Text Generation and Encoding}
\label{sec:text_generation}

\paragraph{Text generation:}
To capture intra-class variability while preserving class semantics, we generate multiple textual descriptions per class (eg: 150, 300, or 450) using a large language model. The descriptions are obtained by systematically varying meaningful class-specific attributes. In action recognition, for example, these attributes include body symmetry, involved body parts, joint-level involvement, movement amplitude, motion plane, execution style, and repetition. Since the considered tasks differ in semantic structure, separate prompts are designed for human action recognition, image classification, and video classification.

For reproducibility, the prompts used for textual description generation are provided separately below.

\begin{enumerate}
    \item \textbf{Prompt for action description generation.} \\
\textbf{Role:} You are a precise biomechanics description engine.  

\noindent\textbf{Action:} \texttt{pickup something}

\noindent\textbf{Attributes to vary (with biomechanical plausibility constraints):}
\begin{itemize}
    \item Side / Symmetry: left, right, both, alternating
    \item Body Part Focus: hand, arm, leg, foot, head, trunk, full body
    \item Joint-Level Emphasis: shoulder, elbow, wrist, hand, hip, knee, ankle, head/neck, trunk
    \item Amplitude: small, medium, large
    \item Tempo / Speed: slow, normal, fast, explosive
    \item Plane / Direction: sagittal, frontal, transverse
    \item Support / Contact Pattern: bilateral, single, alternating, airborne
    \item Style / Effort: relaxed, controlled, casual, stiff, powerful, expressive
    \item Repetition / Duration: single, repeated, continuous, prolonged
    \item Height / Position: low, waist, chest, overhead
\end{itemize}

\noindent\textbf{Task:}
Generate exactly 300 distinct natural-language descriptions of the action
\texttt{pickup something}.

\noindent\textbf{Constraints:}
\begin{itemize}
    \item Each description must incorporate at least three attributes from the list above.
    \item Each description must consist of 3--5 sentences, where each sentence
    describes a sub-action or biomechanical phase of the movement.
    \item All descriptions must be biomechanically realistic, with logically
    ordered action phases and compatible attribute combinations.
    \item Attribute combinations should be systematically varied across samples
    to ensure diversity.
    \item No descriptions may be duplicated.
\end{itemize}

\noindent\textbf{Output format:}
Return an array containing exactly 300 strings.

\item \textbf{Prompt for Image classification task}

\textbf{Role:} You are a precise low-resolution visual appearance description engine specialized for images.

\textbf{Object Class:} frog
\textbf{Attributes to vary (with visual plausibility constraints):}
\begin{itemize}
    \item Viewpoint / Orientation:front | side | rear | angled | top-down
    \item Scale / Object Size:tiny | small | medium
    \item Framing / Visibility:centered | off-center | partially cropped | partially visible
    \item Occlusion Level:none | mild | heavy
    \item Position in Scene:foreground-dominant | background-dominant | blended
    \item Background Context:natural | urban | indoor | cluttered | simple
    \item Lighting Condition:bright | dim | overcast | shadowed
    \item Image Quality / Resolution Artifacts:sharp | slightly blurry | noisy | low-contrast | pixelated
    \item Color / Texture Variation:natural colors | muted colors | high contrast
\end{itemize}

\textbf{Task}:Generate EXACTLY 300 distinct natural-language image descriptions of the object class “frog”.

\textbf{constraints}:
Each description must:
\begin{itemize}
    \item Explicitly mention the object class
    \item Incorporate at least 3 different attributes from the list above
    \item Each description must be 1–2 sentences only
    \item Descriptions must be single-frame only (no temporal progression, no actions)
    \item Ensure low-resolution realism consistent with CIFAR images
    \item Avoid artistic styles (no sketches, cartoons, paintings)
    \item Vary attributes systematically to span intra-class visual variance
    \item Do not repeat descriptions
\end{itemize}

\textbf{Output Format:} Return a JSON array with EXACTLY 300 strings. No explanations. No extra keys.

\item \textbf{Prompt for video classification task}
\textbf{Role:} You are a precise video description engine specialized in visual variation for video classification datasets.

\textbf{Class Label:} <class name>

\textbf{Task:} Generate EXACTLY 150 distinct natural-language descriptions of short video clips depicting the class “<class name>”.

\textbf{Instructions:}\\
\begin{itemize}
    \item Infer the likely visual meaning of “<class name>” from the label itself.
    \item Keep that meaning fixed across all 150 descriptions.
    \item Generate descriptions that reflect realistic intra-class variation while remaining faithful to the same class.    
\end{itemize}

\textbf{Rules:}
\begin{itemize}
    \item Every description must explicitly contain the exact class label “<class name>”.
    \item Every description must describe one short video clip only.
    \item Every description must include at least 4 visible factors chosen from:
    \begin{itemize}
        \item camera angle or viewpoint
        \item framing or crop
        \item subject scale
        \item temporal crop or motion phase visibility
        \item background context
        \item lighting condition
        \item occlusion or partial visibility
    \end{itemize}
    \item Vary the descriptions systematically across clip appearance, motion visibility, camera placement, scene type, and video quality.
    \item Keep the descriptions realistic for natural video datasets.
    \item Keep each description to 1–2 sentences only.
    \item Do not mention attribute names explicitly.
\end{itemize}

\textbf{Output Format:} Return a JSON array with EXACTLY 150 strings. No explanations. No extra keys.

\end{enumerate}

\paragraph{Text encoding.}
Once the action descriptions are generated, each description is encoded using a pretrained sentence-level text encoder. Specifically, we employ different encoders such as BGE \cite{bge}, CLIP \cite{clip}, BERT \cite{bert}.

Each text description is passed through the frozen text encoder, and the final sentence representation is obtained from the output corresponding to the \texttt{[CLS]} token. All text embeddings are $\ell_2$-normalized prior to subsequent processing. These embeddings serve as the input semantic representations to construct the \paperabb{}.

\section{Additional Ablation Studies}
\label{app:ablation_st}
Additional ablation studies are conducted to further analyze the design choices of the proposed \paperabb{}.
\paragraph{Selection of the mapping function $g_\theta$:}
To construct the target semantic space $(\mathcal{C}, \rho_M)$ from the original text-encoder semantic space $(\mathcal{C}, \rho_E)$, \paperabb{} employs a learnable mapping function $g_\theta$. To achieve the proposed separation enhancement induced by the targeted distance transformation, the mapping function must provide sufficient capacity to map semantics from the original space to the target space

In this ablation, the effect of the mapping function is studied by varying the capacity of $g_\theta$, ranging from a single linear projection to multi-layer perceptrons with nonlinear activations. The evaluation is performed using the semantic validation experiments described in \cref{app:semantic-validation}. To isolate the effect of the mapping function, the separation enhancement parameter is fixed to $a = 0.2$ across all experiments.

As shown in \cref{tab:ablation_gtheta}, increasing the capacity of the mapping function $g_\theta$ improves semantic fidelity. A single linear projection provides insufficient flexibility to realize the targeted distance transformation. Introducing nonlinearity and normalization leads to consistent improvements across all semantic validation metrics, including Spearman correlation and KNN overlap, indicating a more faithful mapping to the target distances. Deeper configurations further enhance semantic fidelity, with a moderately expressive nonlinear mapping proving sufficient.

\begin{table}[t]
\centering
\caption{Ablation on the mapping function $g_\theta$. Here, \textbf{L} denotes a linear layer, \textbf{LN} denotes layer normalization, and \textbf{G} denotes the GELU activation. 
Increasing the capacity of $g_\theta$ improves semantic fidelity.}
\label{tab:ablation_gtheta}
\small
\setlength{\tabcolsep}{5pt}
\begin{tabular}{l c ccccc c}
\toprule
\multirow{2}{*}{\textbf{Mapping $g_\theta$}} &
\textbf{Spearman} &
\multicolumn{4}{c}{\textbf{KNN-overlap@k}} &
\textbf{OVR} \\
\cmidrule(lr){3-6}
 &  & 5 & 10 & 50 & 100 &  \\
\midrule
L                                              & 0.93 & 0.57 & 0.63 & 0.83 & 0.95 & 11.4 \\
L $\rightarrow$ G                               & 0.94 & 0.58 & 0.64 & 0.84 & 0.95 & 10.4 \\
L $\rightarrow$ LN $\rightarrow$ G              & 0.98 & 0.60 & 0.67 & 0.91 & 0.98 & 5.40 \\
L $\rightarrow$ LN $\rightarrow$ G $\rightarrow$ L
                                               & 0.99 & 0.96 & 0.96  & 0.99  & 0.99 & 0.63\\
L $\rightarrow$ LN $\rightarrow$ G $\rightarrow$ L $\rightarrow$ LN
                                               & 0.99 & 0.97 & 0.98 & 0.99 & 0.99 & 0.50 \\
L $\rightarrow$ LN $\rightarrow$ G $\rightarrow$ L $\rightarrow$ LN $\rightarrow$ G
                                               & 0.99  & 0.90 & 0.93 & 0.97 & 0.97 & 1.20 \\
L $\rightarrow$ LN $\rightarrow$ G $\rightarrow$ L $\rightarrow$ LN $\rightarrow$ G $\rightarrow$ L
                                               & \textbf{0.99} & \textbf{0.99} & \textbf{0.99} & \textbf{0.99} & \textbf{0.99} & \textbf{0.04} \\
\bottomrule
\end{tabular}
\end{table}

\textbf{Sensitivity to the number of descriptions per class:} The sensitivity of \paperabb{} to the number of LLM-generated descriptions is evaluated on NTU60 using the MAMP motion encoder, as shown in \cref{tab:num_desc}. 
Descriptions per class are varied from 50, 150, 300, to 450, with a separate \paperabb{} constructed for each setting and frozen visual features aligned using the MLP. 
The best performance is obtained with 300 descriptions per class, while results remain relatively stable across settings, indicating that \paperabb{} is not highly sensitive to the number of descriptions.

\textbf{Sensitivity to the text-generating LLM:} As shown in \cref{tab:llm_sens}, LLM sensitivity is evaluated on CIFAR-10 with the MAE encoder by varying the text-generating model while keeping the prompt fixed. ChatGPT-5, Google Gemini~\cite{gemini}, and Claude~\cite{claude} are compared, with Claude and chatGPT-5 achieving the best classification accuracy of 89.2\%. Overall, results remain stable across LLMs, indicating that \paperabb{} is not highly sensitive to the text-generating LLM.

\textbf{Sensitivity to the text encoder:} As shown in \cref{tab:encoder_sens}, text-encoder sensitivity is evaluated on CIFAR-10 with the MAE image encoder using BGE~\cite{bge}, CLIP~\cite{clip}, and BERT~\cite{bert} pretrained text encoders. BGE and CLIP achieve the best performance of 89.2\%, while results remain largely consistent, indicating that \paperabb{} is minimally sensitive to the text encoder choice.

\begin{table}[t]
\centering
\setlength{\tabcolsep}{3pt}

\begin{minipage}[t]{0.32\columnwidth}
\centering
\caption{Impact of the number of text descriptions.}
\label{tab:num_desc}
\begin{tabular}{lc}
\toprule
No. of texts & Top-1 (\%) \\
\midrule
50  & 87.9 \\
150 & 88.0 \\
300 & \textbf{88.2} \\
450 & 88.0 \\
\bottomrule
\end{tabular}
\end{minipage}
\hfill
\begin{minipage}[t]{0.32\columnwidth}
\centering
\caption{Sensitivity of \paperabb{} to the text-generating LLM.}
\label{tab:llm_sens}
\begin{tabular}{lc}
\toprule
\makecell{LLM used to\\generate text} & Top-1 (\%) \\
\midrule
ChatGPT-5 & \textbf{89.2} \\
Claude    & \textbf{89.2}     \\
Gemini    & 89.1     \\
\bottomrule
\end{tabular}
\end{minipage}
\hfill
\begin{minipage}[t]{0.32\columnwidth}
\centering
\caption{Sensitivity of \paperabb{} to the text encoder.}
\label{tab:encoder_sens}
\begin{tabular}{lc}
\toprule
\makecell{Text encoder\\used} & Top-1 (\%) \\
\midrule
BGE  & \textbf{89.2} \\
CLIP & \textbf{89.2} \\
BERT & 89.0 \\
\bottomrule
\end{tabular}
\end{minipage}

\end{table}

\bibliographystyle{plainnat}
\bibliography{references}

@inproceedings{clip,
  title={Learning transferable visual models from natural language supervision},
  author={Radford, Alec and Kim, Jong Wook and Hallacy, Chris and Ramesh, Aditya and Goh, Gabriel and Agarwal, Sandhini and Sastry, Girish and Askell, Amanda and Mishkin, Pamela and Clark, Jack and others},
  booktitle={International conference on machine learning},
  pages={8748--8763},
  year={2021},
  organization={PMLR}
}

@InProceedings{gap,
    author    = {Xiang, Wangmeng and Li, Chao and Zhou, Yuxuan and Wang, Biao and Zhang, Lei},
    title     = {Generative Action Description Prompts for Skeleton-based Action Recognition},
    booktitle = {Proceedings of the IEEE/CVF International Conference on Computer Vision (ICCV)},
    month     = {October},
    year      = {2023},
    pages     = {10276-10285}
}

@article{crossglg,
  title={CrossGLG: LLM Guides One-shot Skeleton-based 3D Action Recognition in a Cross-level Manner},
  author={Tingbing Yan and Wenzheng Zeng and Yang Xiao and Xingyu Tong and Bo Tan and Zhiwen Fang and Zhiguo Cao and Joey Tianyi Zhou},
  journal={ArXiv},
  year={2024},
  volume={abs/2403.10082},
  url={https://api.semanticscholar.org/CorpusID:268509807}
}

@InProceedings{purls,
    author    = {Zhu, Anqi and Ke, Qiuhong and Gong, Mingming and Bailey, James},
    title     = {Part-aware Unified Representation of Language and Skeleton for Zero-shot Action Recognition},
    booktitle = {Proceedings of the IEEE/CVF Conference on Computer Vision and Pattern Recognition (CVPR)},
    month     = {June},
    year      = {2024},
    pages     = {18761-18770}
}

@inproceedings{bge,
author = {Xiao, Shitao and Liu, Zheng and Zhang, Peitian and Muennighoff, Niklas and Lian, Defu and Nie, Jian-Yun},
title = {C-Pack: Packed Resources For General Chinese Embeddings},
year = {2024},
isbn = {9798400704314},
publisher = {Association for Computing Machinery},
address = {New York, NY, USA},
url = {https://doi.org/10.1145/3626772.3657878},
doi = {10.1145/3626772.3657878},
booktitle = {Proceedings of the 47th International ACM SIGIR Conference on Research and Development in Information Retrieval},
pages = {641–649},
numpages = {9},
location = {Washington DC, USA},
series = {SIGIR '24}
}

@InProceedings{ntu60,
author = {Shahroudy, Amir and Liu, Jun and Ng, Tian-Tsong and Wang, Gang},
title = {NTU RGB+D: A Large Scale Dataset for 3D Human Activity Analysis},
booktitle = {Proceedings of the IEEE Conference on Computer Vision and Pattern Recognition (CVPR)},
month = {June},
year = {2016}
}

@article{ntu120,
author = {Liu, Jun and Shahroudy, Amir and Perez, Mauricio and Wang, Gang and Duan, Ling-Yu and Kot, Alex C.},
title = {NTU RGB+D 120: A Large-Scale Benchmark for 3D Human Activity Understanding},
year = {2020},
issue_date = {Oct. 2020},
publisher = {IEEE Computer Society},
address = {USA},
volume = {42},
number = {10},
issn = {0162-8828},
url = {https://doi.org/10.1109/TPAMI.2019.2916873},
doi = {10.1109/TPAMI.2019.2916873},
journal = {IEEE Trans. Pattern Anal. Mach. Intell.},
month = oct,
pages = {2684–2701},
numpages = {18}
}

@article{actionclip,
  title={ActionCLIP: A New Paradigm for Video Action Recognition},
  author={Mengmeng Wang and Jiazheng Xing and Yong Liu},
  journal={ArXiv},
  year={2021},
  volume={abs/2109.08472},
  url={https://api.semanticscholar.org/CorpusID:237563206}
}

@article{gemini,
  title={Gemini: a family of highly capable multimodal models},
  author={Team, Gemini and Anil, Rohan and Borgeaud, Sebastian and Alayrac, Jean-Baptiste and Yu, Jiahui and Soricut, Radu and Schalkwyk, Johan and Dai, Andrew M and Hauth, Anja and Millican, Katie and others},
  journal={arXiv preprint arXiv:2312.11805},
  year={2023}
}

@misc{claude,
  title={The Claude 3 Model Family: Opus, Sonnet, Haiku},
  author={Anthropic},
  year={2024},
  url={https://api.semanticscholar.org/CorpusID:268232499}
}

@inproceedings{bert,
  author = {Devlin, Jacob and Chang, Ming-Wei and Lee, Kenton and Toutanova, Kristina},
  booktitle = {NAACL-HLT (1)},
  editor = {Burstein, Jill and Doran, Christy and Solorio, Thamar},
  ee = {https://www.aclweb.org/anthology/N19-1423/},
  isbn = {978-1-950737-13-0},
  pages = {4171-4186},
  publisher = {Association for Computational Linguistics},
  title = {BERT: Pre-training of Deep Bidirectional Transformers for Language Understanding.},
  url = {http://dblp.uni-trier.de/db/conf/naacl/naacl2019-1.html#DevlinCLT19},
  year = 2019
}

@article{neuron,
  title={Neuron: Learning Context-Aware Evolving Representations for Zero-Shot Skeleton Action Recognition},
  author={Yang Chen and Jingcai Guo and Song Guo and Dacheng Tao},
  journal={2025 IEEE/CVF Conference on Computer Vision and Pattern Recognition (CVPR)},
  year={2024},
  pages={8721-8730},
  url={https://api.semanticscholar.org/CorpusID:274131291}
}

@misc{FSVAE,
author = {Wu, Wenhan and Guo, Zhishuai and Chen, Chen and Xue, Hongfei and Lu, Aidong},
year = {2025},
month = {06},
pages = {},
title = {Frequency-Semantic Enhanced Variational Autoencoder for Zero-Shot Skeleton-based Action Recognition},
doi = {10.48550/arXiv.2506.22179}
}

@article{knn,
  title={Embedding Comparator: Visualizing Differences in Global Structure and Local Neighborhoods via Small Multiples},
  author={Angie Boggust and Brandon Carter and Arvind Satyanarayan},
  journal={Proceedings of the 27th International Conference on Intelligent User Interfaces},
  year={2019},
  url={https://api.semanticscholar.org/CorpusID:209140747}
}

@article{kendal,
  title={Cross-Care: Assessing the Healthcare Implications of Pre-training Data on Language Model Bias},
  author={Shan Chen and Jack Gallifant and Mingye Gao and Pedro Moreira and Nikolaj Munch and Ajay Muthukkumar and Arvind Rajan and Jaya Kolluri and Amelia Fiske and Janna Hastings and Hugo J. W. L. Aerts and Brian Anthony and Leo Anthony Celi and William G. La Cava and Danielle S. Bitterman},
  journal={Advances in neural information processing systems},
  year={2024},
  volume={37 D-ampB},
  pages={
          23756-23795
        },
  url={https://api.semanticscholar.org/CorpusID:269635575}
}

@inproceedings{spearman,
  author       = {Frances Ding and
                  Jean{-}Stanislas Denain and
                  Jacob Steinhardt},
  editor       = {Marc'Aurelio Ranzato and
                  Alina Beygelzimer and
                  Yann N. Dauphin and
                  Percy Liang and
                  Jennifer Wortman Vaughan},
  title        = {Grounding Representation Similarity Through Statistical Testing},
  booktitle    = {Advances in Neural Information Processing Systems 34: Annual Conference
                  on Neural Information Processing Systems 2021, NeurIPS 2021, December
                  6-14, 2021, virtual},
  pages        = {1556--1568},
  year         = {2021},
  url          = {https://proceedings.neurips.cc/paper/2021/hash/0c0bf917c7942b5a08df71f9da626f97-Abstract.html},
  bibsource    = {dblp computer science bibliography, https://dblp.org}
}

@inproceedings{ovr,
title = "Generalization Error Bound for Hyperbolic Ordinal Embedding",
author = "Atsushi Suzuki and Atsushi Nitanda and Wang, \{Jing Jing\} and Linchuan Xu and Kenji Yamanishi and Marc Cavazza",
year = "2021",
language = "English",
pages = "10011--10021",
booktitle = "Proceedings of the 38th International Conference on Machine Learning",
}

@TECHREPORT{cifar10,
    author = {Alex Krizhevsky},
    title = {Learning multiple layers of features from tiny images},
    institution = {University of Toronto},
    year = {2009},
}

@article{lda,
  title={The use of multiple measurements in taxonomic problems},
  author={Fisher, Ronald A},
  journal={Annals of eugenics},
  volume={7},
  number={2},
  pages={179--188},
  year={1936},
  publisher={Wiley Online Library}
}

@article{LFDA,
  title={Dimensionality Reduction of Multimodal Labeled Data by Local Fisher Discriminant Analysis},
  author={Masashi Sugiyama},
  journal={J. Mach. Learn. Res.},
  year={2007},
  volume={8},
  pages={1027-1061},
  url={https://api.semanticscholar.org/CorpusID:9678351}
}

@article{MFA,
  title={Graph embedding and extensions: A general framework for dimensionality reduction},
  author={Yan, Shuicheng and Xu, Dong and Zhang, Benyu and Zhang, Hong-Jiang and Yang, Qiang and Lin, Stephen},
  journal={IEEE transactions on pattern analysis and machine intelligence},
  volume={29},
  number={1},
  pages={40--51},
  year={2006},
  publisher={IEEE}
}

@inproceedings{graphlda,
author = {Cai, Deng and He, Xiaofei and Zhou, Kun and Han, Jiawei and Bao, Hujun},
title = {Locality sensitive discriminant analysis},
year = {2007},
publisher = {Morgan Kaufmann Publishers Inc.},
address = {San Francisco, CA, USA},
booktitle = {Proceedings of the 20th International Joint Conference on Artifical Intelligence},
pages = {708–713},
numpages = {6},
location = {Hyderabad, India},
series = {IJCAI'07}
}

@ARTICLE{isomap,
  author={Xin Geng and De-Chuan Zhan and Zhi-Hua Zhou},
  journal={IEEE Transactions on Systems, Man, and Cybernetics, Part B (Cybernetics)}, 
  title={Supervised nonlinear dimensionality reduction for visualization and classification}, 
  year={2005},
  volume={35},
  number={6},
  pages={1098-1107},
  doi={10.1109/TSMCB.2005.850151}}

@article{mds,
  title={Multidimensional scaling by optimizing goodness of fit to a nonmetric hypothesis},
  author={Joseph B. Kruskal},
  journal={Psychometrika},
  year={1964},
  volume={29},
  pages={1-27},
  url={https://api.semanticscholar.org/CorpusID:48165675}
}

@article{umap,
  title={Umap: Uniform manifold approximation and projection for dimension reduction},
  author={McInnes, Leland and Healy, John and Melville, James},
  journal={arXiv preprint arXiv:1802.03426},
  year={2018}
}

@inproceedings{Ethayarajh2019HowCA,
  title={How Contextual are Contextualized Word Representations? Comparing the Geometry of BERT, ELMo, and GPT-2 Embeddings},
  author={Kawin Ethayarajh},
  booktitle={Conference on Empirical Methods in Natural Language Processing},
  year={2019},
  url={https://api.semanticscholar.org/CorpusID:202120592}
}

@inproceedings{mae,
  title={Masked autoencoders are scalable vision learners},
  author={He, Kaiming and Chen, Xinlei and Xie, Saining and Li, Yanghao and Doll{\'a}r, Piotr and Girshick, Ross},
  booktitle={Proceedings of the IEEE/CVF conference on computer vision and pattern recognition},
  pages={16000--16009},
  year={2022}
}

@article{foundskel,
  title={Foundation model for skeleton-based human action understanding},
  author={Wang, Hongsong and Weng, Wanjiang and Wang, Junbo and Zhao, Fang and Xie, Guo-Sen and Geng, Xin and Wang, Liang},
  journal={IEEE Transactions on Pattern Analysis and Machine Intelligence},
  year={2025},
  publisher={IEEE}
}

@inproceedings{mamp,
  title={Masked motion predictors are strong 3d action representation learners},
  author={Mao, Yunyao and Deng, Jiajun and Zhou, Wengang and Fang, Yao and Ouyang, Wanli and Li, Houqiang},
  booktitle={Proceedings of the IEEE/CVF International Conference on Computer Vision},
  pages={10181--10191},
  year={2023}
}

@article{coop,
  title={Learning to prompt for vision-language models},
  author={Zhou, Kaiyang and Yang, Jingkang and Loy, Chen Change and Liu, Ziwei},
  journal={International Journal of Computer Vision},
  volume={130},
  number={9},
  pages={2337--2348},
  year={2022},
  publisher={Springer}
}

@inproceedings{cocoop,
  title={Conditional prompt learning for vision-language models},
  author={Zhou, Kaiyang and Yang, Jingkang and Loy, Chen Change and Liu, Ziwei},
  booktitle={Proceedings of the IEEE/CVF conference on computer vision and pattern recognition},
  pages={16816--16825},
  year={2022}
}

@inproceedings{maple,
  title={Maple: Multi-modal prompt learning},
  author={Khattak, Muhammad Uzair and Rasheed, Hanoona and Maaz, Muhammad and Khan, Salman and Khan, Fahad Shahbaz},
  booktitle={Proceedings of the IEEE/CVF conference on computer vision and pattern recognition},
  pages={19113--19122},
  year={2023}
}

@inproceedings{promptsrc,
  title={Self-regulating prompts: Foundational model adaptation without forgetting},
  author={Khattak, Muhammad Uzair and Wasim, Syed Talal and Naseer, Muzammal and Khan, Salman and Yang, Ming-Hsuan and Khan, Fahad Shahbaz},
  booktitle={Proceedings of the IEEE/CVF international conference on computer vision},
  pages={15190--15200},
  year={2023}
}

@inproceedings{tip,
  title={Tip-adapter: Training-free adaption of clip for few-shot classification},
  author={Zhang, Renrui and Zhang, Wei and Fang, Rongyao and Gao, Peng and Li, Kunchang and Dai, Jifeng and Qiao, Yu and Li, Hongsheng},
  booktitle={European conference on computer vision},
  pages={493--510},
  year={2022},
  organization={Springer}
}

@article{watt,
  title={WATT: Weight average test time adaptation of CLIP},
  author={Osowiechi, David and Noori, Mehrdad and Vargas Hakim, Gustavo and Yazdanpanah, Moslem and Bahri, Ali and Cheraghalikhani, Milad and Dastani, Sahar and Beizaee, Farzad and Ayed, Ismail and Desrosiers, Christian},
  journal={Advances in neural information processing systems},
  volume={37},
  pages={48015--48044},
  year={2024}
}

@article{karlgren2021semantics,
  title={Semantics in high-dimensional space},
  author={Karlgren, Jussi and Kanerva, Pentti},
  journal={Frontiers in artificial intelligence},
  volume={4},
  pages={698809},
  year={2021},
  publisher={Frontiers Media SA}
}

@article{schall2025comprehensive,
  title={A comprehensive approach to improving CLIP-based image retrieval while maintaining joint-embedding alignment},
  author={Schall, Konstantin and Barthel, Kai Uwe and Hezel, Nico and Moelle, Andre},
  journal={Information Systems},
  volume={134},
  pages={102581},
  year={2025},
  publisher={Elsevier}
}

@misc{cub200,
doi = {10.21227/fg6w-vh29},
url = {https://dx.doi.org/10.21227/fg6w-vh29},
author = {Catherine Wah and Steve Branson and Peter Welinder and Pietro Perona and Serge Belongie},
publisher = {IEEE Dataport},
title = {Caltech-UCSD Birds-200-2011},
year = {2025} }

@misc{dinov3,
  title={{DINOv3}},
  author={Sim{\'e}oni, Oriane and Vo, Huy V. and Seitzer, Maximilian and Baldassarre, Federico and Oquab, Maxime and Jose, Cijo and Khalidov, Vasil and Szafraniec, Marc and Yi, Seungeun and Ramamonjisoa, Micha{\"e}l and Massa, Francisco and Haziza, Daniel and Wehrstedt, Luca and Wang, Jianyuan and Darcet, Timoth{\'e}e and Moutakanni, Th{\'e}o and Sentana, Leonel and Roberts, Claire and Vedaldi, Andrea and Tolan, Jamie and Brandt, John and Couprie, Camille and Mairal, Julien and J{\'e}gou, Herv{\'e} and Labatut, Patrick and Bojanowski, Piotr},
  year={2025},
  eprint={2508.10104},
  archivePrefix={arXiv},
  primaryClass={cs.CV},
  url={https://arxiv.org/abs/2508.10104},
}

@inproceedings{videomae,
  title={Videomae v2: Scaling video masked autoencoders with dual masking},
  author={Wang, Limin and Huang, Bingkun and Zhao, Zhiyu and Tong, Zhan and He, Yinan and Wang, Yi and Wang, Yali and Qiao, Yu},
  booktitle={Proceedings of the IEEE/CVF conference on computer vision and pattern recognition},
  pages={14549--14560},
  year={2023}
}

@article{kulis2013metric,
  title={Metric learning: A survey},
  author={Kulis, Brian},
  journal={Foundations and Trends{\textregistered} in Machine Learning},
  volume={5},
  number={4},
  pages={287--364},
  year={2013},
  publisher={Emerald Publishing Limited}
}

@article{supcon,
  title={Supervised contrastive learning},
  author={Khosla, Prannay and Teterwak, Piotr and Wang, Chen and Sarna, Aaron and Tian, Yonglong and Isola, Phillip and Maschinot, Aaron and Liu, Ce and Krishnan, Dilip},
  journal={Advances in neural information processing systems},
  volume={33},
  pages={18661--18673},
  year={2020}
}

@inproceedings{msloss,
  title={Multi-similarity loss with general pair weighting for deep metric learning},
  author={Wang, Xun and Han, Xintong and Huang, Weilin and Dong, Dengke and Scott, Matthew R},
  booktitle={Proceedings of the IEEE/CVF conference on computer vision and pattern recognition},
  pages={5022--5030},
  year={2019}
}

@inproceedings{circleloss,
  title={Circle loss: A unified perspective of pair similarity optimization},
  author={Sun, Yifan and Cheng, Changmao and Zhang, Yuhan and Zhang, Chi and Zheng, Liang and Wang, Zhongdao and Wei, Yichen},
  booktitle={Proceedings of the IEEE/CVF conference on computer vision and pattern recognition},
  pages={6398--6407},
  year={2020}
}

@inproceedings{proxyanchor,
  title={Proxy Anchor Loss for Deep Metric Learning},
  author={Kim, Sungyeon and Kim, Dongwon and Cho, Minsu and Kwak, Suha},
  booktitle={Proceedings of the IEEE/CVF Conference on Computer Vision and Pattern Recognition},
  year={2020}
}

@inproceedings{arcface,
  title={Arcface: Additive angular margin loss for deep face recognition},
  author={Deng, Jiankang and Guo, Jia and Xue, Niannan and Zafeiriou, Stefanos},
  booktitle={Proceedings of the IEEE/CVF conference on computer vision and pattern recognition},
  pages={4690--4699},
  year={2019}
}

@article{tsne,
  title={Visualizing data using t-SNE.},
  author={Van der Maaten, Laurens and Hinton, Geoffrey},
  journal={Journal of machine learning research},
  volume={9},
  number={11},
  year={2008}
}

@InProceedings{oxfordiit,
  author       = "Omkar M. Parkhi and Andrea Vedaldi and Andrew Zisserman and C. V. Jawahar",
  title        = "Cats and Dogs",
  booktitle    = "IEEE Conference on Computer Vision and Pattern Recognition",
  year         = "2012",
}

@inproceedings{eval_metric,
  title={Cross-modal retrieval with correspondence autoencoder},
  author={Feng, Fangxiang and Wang, Xiaojie and Li, Ruifan},
  booktitle={Proceedings of the 22nd ACM international conference on Multimedia},
  pages={7--16},
  year={2014}
}

@inproceedings{hmdb,
  title={HMDB: a large video database for human motion recognition},
  author={Kuehne, Hildegard and Jhuang, Hueihan and Garrote, Est{\'\i}baliz and Poggio, Tomaso and Serre, Thomas},
  booktitle={2011 International conference on computer vision},
  pages={2556--2563},
  year={2011},
  organization={IEEE}
}

@inproceedings{arid,
  title={Arid: A new dataset for recognizing action in the dark},
  author={Xu, Yuecong and Yang, Jianfei and Cao, Haozhi and Mao, Kezhi and Yin, Jianxiong and See, Simon},
  booktitle={International workshop on deep learning for human activity recognition},
  pages={70--84},
  year={2021},
  organization={Springer}
}

@inproceedings{thoker2025smile,
  title={Smile: Infusing spatial and motion semantics in masked video learning},
  author={Thoker, Fida Mohammad and Jiang, Letian and Zhao, Chen and Ghanem, Bernard},
  booktitle={Proceedings of the Computer Vision and Pattern Recognition Conference},
  pages={8438--8449},
  year={2025}
}

\end{document}